\documentclass{article} 
\usepackage{arxiv_iclr2027,times}

\usepackage{amsmath,amsfonts,bm}

\def\eqref#1{equation~\ref{#1}}

\def\1{\bm{1}}

\DeclareMathAlphabet{\mathsfit}{\encodingdefault}{\sfdefault}{m}{sl}
\SetMathAlphabet{\mathsfit}{bold}{\encodingdefault}{\sfdefault}{bx}{n}

\usepackage{hyperref}
\usepackage{url}
\usepackage{algorithm}
\usepackage{algorithmic}
\usepackage{graphicx}
\usepackage{subcaption}
\usepackage{wrapfig}
\usepackage{booktabs}
\usepackage{multirow}
\usepackage{placeins}
\title{RoboDrop: Curating VLA Post-Training Data via Local Gradient Compatibility
}

\author{Runze Xu$^{1,2}$, Yuanfan Xu$^{2}$, Cuijie Xu$^{1}$, Shuang Dai$^{1}$, Yining Li$^{1}$ \\
Yu Wang$^{1}$, Jincheng Yu$^{1*}$ \\[2pt]
{\normalfont $^{1}$Tsinghua University \qquad $^{2}$Striding AI}}

\iclrfinalcopy
\begin{document}

\maketitle
\begin{abstract}
Vision--language--action (VLA) models acquire broad generalization through large-scale pretraining, yet adapting them to a new task and robot embodiment still requires post-training on newly collected data. Unlike pretraining, post-training targets task- and embodiment-specific adaptation, making it particularly sensitive to data quality. In practice, collected robot datasets often contain heterogeneous errors, including execution mistakes, sensor drift, and timestamp misalignment, which can impair post-training and policy performance. Manual inspection is costly, while existing data-cleaning methods are typically tailored to particular corruption types. 
To address these challenges, we introduce \textsc{RoboDrop}, a data-curation framework that audits supervision using local gradient compatibility measured along the training trajectory as a proxy for its effect on post-training performance. During a one-epoch warm-up run, RoboDrop scores each candidate sample online by comparing its gradient with those of task-semantic and visually matched validation samples. The resulting sample scores are aggregated at the episode level, and a simple automatic post-processing rule converts them into filtering decisions. In real-world data curation, RoboDrop substantially reduces the manual inspection burden by requiring verification of only a small reference set.
We evaluate RoboDrop on controlled observation--action corruptions, naturally suboptimal demonstrations in simulation, and real-robot datasets containing non-expert collection errors. Across these settings, RoboDrop more accurately distinguishes unreliable demonstrations than prior methods, while post-training on the curated data consistently yields stronger downstream policies, with average real-robot rollout success rising from $35.0\%$ to
$67.5\%$. These results establish training-trajectory-aware, context-conditioned supervision auditing as an effective approach to robust VLA post-training. 
Code is available at \url{https://github.com/shuiyigong/RoboDrop}.
\end{abstract}

\setlength{\abovedisplayskip}{5pt plus 1pt minus 2pt}
\setlength{\belowdisplayskip}{5pt plus 1pt minus 2pt}
\setlength{\abovedisplayshortskip}{3pt plus 1pt minus 1pt}
\setlength{\belowdisplayshortskip}{4pt plus 1pt minus 1pt}
\setlength{\jot}{2pt}
\captionsetup{skip=3pt}

\section{Introduction}
Large-scale vision--language--action (VLA) models have demonstrated broad generalization across diverse robotic tasks and environments~\citep{black2024pi0,black2025pi05,nvidia2025gr00tn1openfoundation,liu2024rdt,brohan2023rt,kim2024openvla,team2024octo}. However, achieving reliable performance on a new task and robot embodiment still requires post-training on task-specific demonstrations collected in the target environment~\citep{kim2025fine,o2024open}. Ensuring the quality of these post-training data is challenging, as real-world
collection is susceptible to imperfect teleoperation, execution mistakes, sensor drift, and temporal misalignment between observations and actions~\citep{li2026wall,li2025gr}. These imperfections need not make an entire trajectory unsuccessful. Instead, they often induce locally unreliable observation--action supervision at particular task states, producing policy updates that
conflict with the desired behavior and ultimately reduce downstream task success. Identifying such episodes by manual inspection is prohibitively expensive.

Existing robot data-selection methods address related but distinct challenges. Simple proxies, such as demonstration length or trajectory-level outcomes, efficiently identify clearly inefficient or failed trajectories, but these coarse signals miss localized sensing and temporal errors within otherwise successful ones. Retrieval-based approaches select data according to visual, motion, or behavioral similarity to a target task~\citep{memmel2025strap,du2023behavior,lin2024flowretrieval}; similarity identifies relevant data, but does not determine whether the action paired with a particular observation provides reliable supervision. Performance-aware approaches, such as DataMIL~\citep{dass2026datamil}, estimate the utility of data for improving a target policy and enable task-aware selection from large, heterogeneous data collections. In contrast, we focus on identifying locally unreliable supervision within task-relevant post-training demonstrations. This calls for a fine-grained, context-conditioned assessment of each observation--action pair's contribution to post-training performance.

Gradient-based data attribution provides a useful perspective for this problem. Under a first-order approximation of validation-loss change, the contribution of a training example can be characterized by its gradient alignment with held-out validation behavior~\citep{koh2017understanding,pruthi2020estimating,wang2025data}. However, robot manipulation data are highly heterogeneous: a single task can contain visually and behaviorally distinct reaching, grasping, transport, and placement stages, while visually similar observations can require different actions under different language instructions. Reliable validation behavior should therefore be defined locally, using samples that are compatible with both the task semantics and interaction state of the candidate. Moreover, for flow-matching~\citep{lipman2022flow} VLAs, gradient estimates evolve throughout training and depend on stochastic noising variables, making a single static comparison insufficiently robust.

We introduce \textbf{RoboDrop}, a training-trajectory-aware framework for automated data curation based on \emph{task-semantic and visual-context-conditioned gradient compatibility}. Under this local view, reliable supervision should induce an update compatible with the update supported by validation samples from the same task semantics and interaction state. During a one-epoch warm-up run, RoboDrop scores each candidate sample once, at the optimization step when it is encountered, against the most recently refreshed local validation reference. It then aggregates these online sample-level scores at the trajectory level to preserve the temporal integrity of demonstrations. In practice, the episode scores can be used with a standard fixed-ratio filtering rule. For settings where the contamination rate is unknown, we additionally provide an optional BIC-based post-processing procedure that adaptively converts the episode scores into filtering decisions.

We evaluate RoboDrop across three complementary sources of imperfect supervision: controlled synthetic corruptions that explicitly disrupt observation--action correspondence, naturally suboptimal demonstrations collected by less-skilled operators in simulation, and real-robot datasets containing execution mistakes from non-expert data collection. RoboDrop consistently provides stronger episode-level discrimination than prior methods across multiple VLA backbones, achieving a mean AUROC of $88.33\%$ for distinguishing proficient from suboptimal demonstrations across four simulated manipulation
tasks. On real-robot data, RoboDrop automatically determines the filtering
budget without access to the contamination rate, and post-training on the
curated data improves rollout success from $35.0\%$ to $67.5\%$. Our contributions are as follows:
\begin{itemize}

\item We introduce a \emph{context-conditioned local gradient reference} for auditing VLA post-training supervision. By constructing the reference from clean validation samples matched by task semantics and visual interaction context, it provides a candidate-specific measure of whether each sample induces a desirable policy update.
\item We develop a \emph{training-trajectory-aware online scoring procedure} that evaluates candidate samples as training progresses. It captures fine-grained variations in gradient compatibility during optimization, producing quality estimates that reflect evolving training dynamics.
\item We validate RoboDrop on controlled supervision corruptions, naturally suboptimal simulated demonstrations, and real-robot collection errors, demonstrating improved filtering quality and downstream policy success across multiple VLA backbones.
\end{itemize}

\section{Related Work}

\paragraph{Robot Data Curation and Selection.}
Prior work has explored robot-data selection at multiple granularities~\citep{hejna2024remixoptimizingdatamixtures}. Trajectory-level methods assess demonstration quality using mutual information~\citep{hejna2025robotdatacurationmutual} or manually specified quality signals~\citep{yuan2026qwen}, while retrieval-based approaches select data that are visually, behaviorally, or temporally related to a target task~\citep{memmel2025strap,du2023behavior,lin2024flowretrieval,kumar2025collage}. Rollout-based methods leverage policy execution feedback, either by learning success--failure classifiers to filter demonstrations~\citep{chen2025curating} or by estimating their influence on closed-loop returns~\citep{agia2025cupid}. Fine-grained curation methods further identify suboptimal transitions through task-progress estimation or remove redundant state--action patterns~\citep{zhang2025scizor}. DataMIL estimates data utility for a target policy to select beneficial trajectories from heterogeneous prior datasets~\citep{dass2026datamil,engstrom2024dsdmmodelawaredatasetselection}. QoQ applies gradient-based influence estimation to robot curation by scoring transitions at a fixed checkpoint against individual validation samples~\citep{lee2026quality}. However, static or task-level quality signals can be insufficient for multi-stage robot behavior: supervision that is appropriate for one interaction state may be incompatible with another, and its utility may change as the policy is post-trained. RoboDrop addresses these challenges jointly by auditing supervision relative to context-matched clean behavior along the post-training trajectory.

\paragraph{Gradient-Based Data Attribution.}
Data valuation methods quantify the contribution of training examples to model behavior or validation performance~\citep{Hammoudeh_2024}. Data Shapley provides an axiomatic formulation based on expected marginal utility over data subsets, but typically requires repeated retraining and is consequently expensive~\citep{ghorbani2019data}. Influence functions and gradient-based estimators provide tractable approximations to a sample's effect on validation loss~\citep{koh2017understanding, pruthi2020estimating, park2023trak}. In-Run Data Shapley further accumulates first-order training--validation gradient interactions over a single optimization trajectory~\citep{wang2025data}. Although gradient-based attribution is broadly applicable, recent formulations have primarily been developed for data valuation in language-model training. Robot demonstrations exhibit a different structure: an episode traverses heterogeneous interaction stages, and supervision may be unreliable only at particular states. A single global validation objective can therefore obscure whether a candidate action is appropriate for its specific task and interaction context. RoboDrop adapts training-trajectory attribution to robot post-training by constructing context-conditioned validation references, thereby assessing whether local supervision supports clean behavior at a comparable task state.

\section{Preliminaries}
\label{sec:preliminaries}

We study the post-training of a pretrained VLA policy $\pi_{\theta_0}$ on a collection of task-specific candidate episodes $\mathcal{E}_{\mathrm{cand}}=\{\tau_i\}_{i=1}^{N}$, where $\tau_i=\{z_{i,r}\}_{r=1}^{L_i}$ and $z_{i,r}=(o_{i,r},\ell_i,A_{i,r})$ comprises a multimodal observation, a language instruction, and an action chunk.

We consider flow-matching VLA post-training. For a sample
$z=(o,\ell,A)$, we draw a Gaussian noise vector
$A^0\sim\mathcal{N}(0,I)$ and a flow time
$\lambda\sim\mathcal{U}(0,1)$, forming the interpolated action
$A^\lambda=(1-\lambda)A^0+\lambda A$. The velocity field
$v_\theta(A^\lambda,o,\ell,\lambda)$ is trained to transport $A^0$ toward the
demonstrated action chunk $A$. For a set of
episodes $\mathcal{S}$, post-training minimizes the flow-matching objective
\begin{equation}
    \mathcal{L}_{\mathrm{FM}}(\theta;\mathcal{S})
    =
    \mathbb{E}_{z\sim\mathcal{Z}(\mathcal{S}),\,
    A^0\sim\mathcal{N}(0,I),\,
    \lambda\sim\mathcal{U}(0,1)}
    \left[
        \left\|
        v_\theta(A^\lambda,o,\ell,\lambda)
        -
        (A-A^0)
        \right\|_2^2
    \right],
    \label{eq:flow_matching_loss}
\end{equation}
where $\mathcal{Z}(\mathcal{S})$ denotes the set of samples contained in
episodes $\mathcal{S}$, and
$\theta^\star(\mathcal{S})$ denotes the parameters obtained by minimizing
$\mathcal{L}_{\mathrm{FM}}(\theta;\mathcal{S})$ from initialization
$\theta_0$.

Although all candidate episodes are collected for the target setting, they may
contain locally unreliable observation--action supervision due to imperfect
teleoperation or sensing errors. The ideal curation
objective is to select the episodes that maximize the downstream task success
rate of the resulting post-trained policy:
\begin{equation}
    \mathcal{S}^{\star}
    \in
    \arg\max_{\mathcal{S}\subseteq\mathcal{E}_{\mathrm{cand}}}
    \operatorname{Success}\!\left(
        \pi_{\theta^\star(\mathcal{S})}
    \right).
    \label{eq:ideal_curation_objective}
\end{equation}

\begin{figure*}[!t]
    \centering
    \begin{subfigure}[t]{0.56\textwidth}
        \centering
        \includegraphics[width=\linewidth]{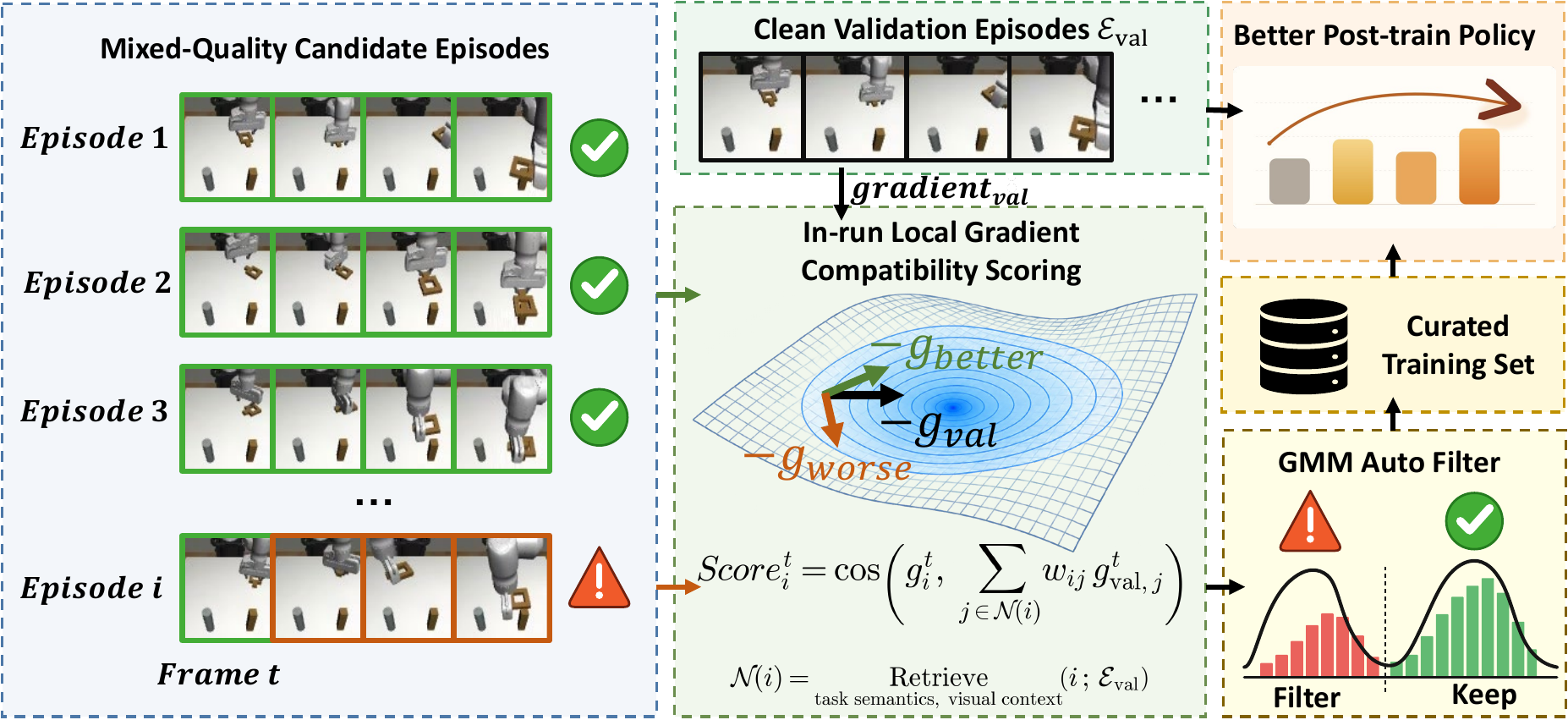}
        \caption{RoboDrop pipeline.}
        \label{fig:overview}
    \end{subfigure}\hfill
    \begin{subfigure}[t]{0.42\textwidth}
        \centering
        \includegraphics[width=\linewidth]{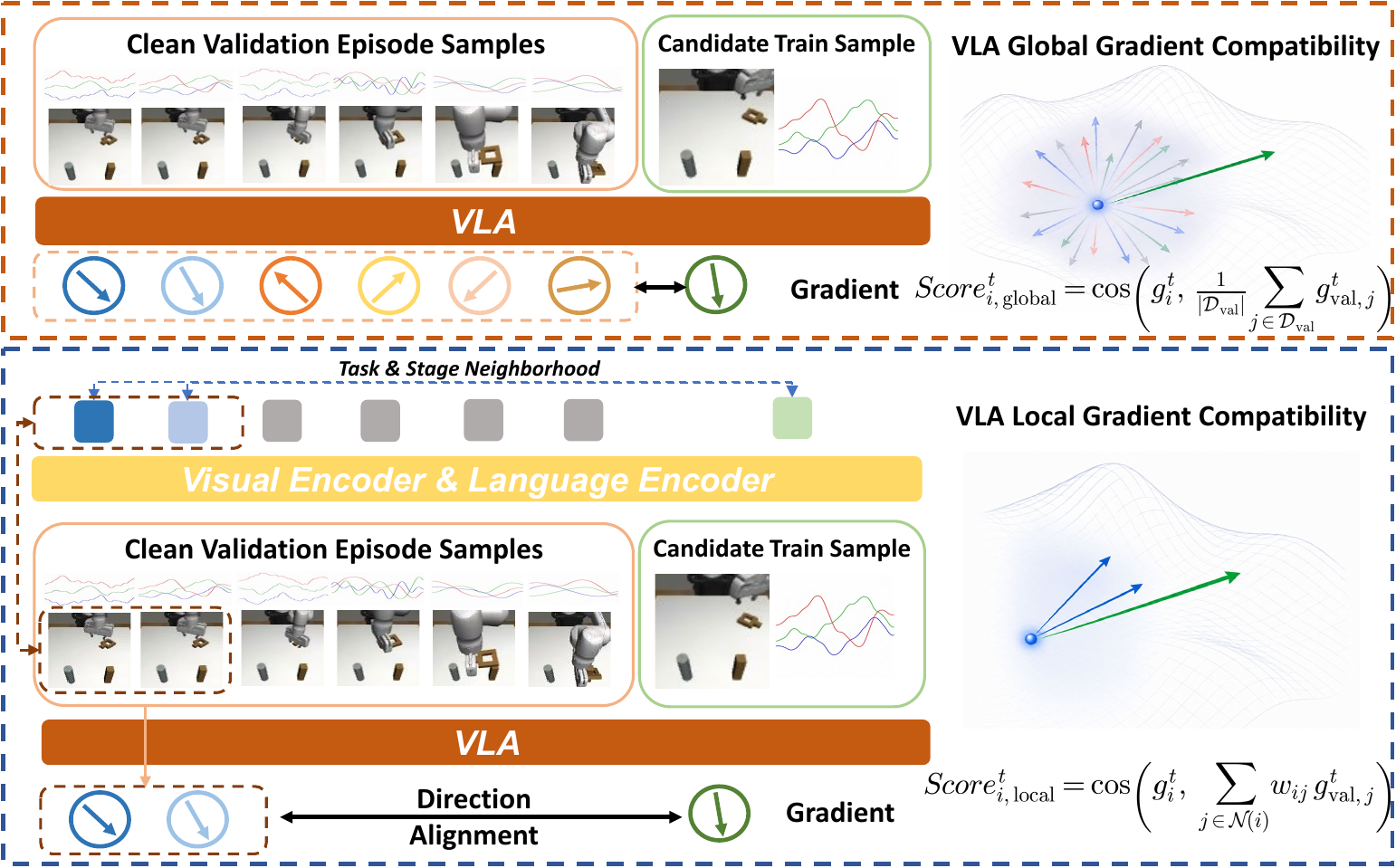}
        \caption{Global versus local reference gradients.}
        \label{fig:global_local}
    \end{subfigure}
    \caption{RoboDrop overview and global-versus-local reference construction.}
    \label{fig:method}
\end{figure*}

\section{RoboDrop}
\label{sec:method}

\subsection{Overview}
\label{sec:method_overview}
Figure~\ref{fig:method}\subref{fig:overview} summarizes RoboDrop. Given
mixed-quality candidate episodes $\mathcal{E}_{\mathrm{cand}}$ and a small clean reference set $\mathcal{E}_{\mathrm{val}}$, RoboDrop performs a one-epoch warm-up run and scores each candidate sample online. Each score measures the sample's gradient compatibility with clean validation behavior from similar context, thereby capturing supervision quality along the evolving post-training trajectory. RoboDrop then aggregates the sample-level scores by episode. When the contamination rate is unknown, besides fixed-ratio filtering, we provide an optional BIC-based procedure that recommends a removal ratio based on the episode-score distribution. Finally, the warm-up policy is discarded, and a fresh pretrained policy is post-trained on the retained episodes together with $\mathcal{E}_{\mathrm{val}}$. 

\subsection{From Validation Loss to Gradient Compatibility}
\label{sec:gradient_proxy}

Directly optimizing Equation~\ref{eq:ideal_curation_objective} is intractable. Rollout success is non-differentiable with respect to data selection, and evaluating a candidate subset requires post-training a policy and deploying it for environment rollouts. Repeating this procedure over possible subsets is prohibitively expensive. We therefore seek a differentiable proxy that can be evaluated within a single training run.

We assess the policy using its flow-matching loss on the small collection of clean target-task episodes $\mathcal{E}_{\mathrm{val}}$, consistent with the use of held-out imitation loss as a surrogate for policy quality in model-aware data curation~\citep{dass2026datamil}. Let $\mathcal{D}_{\mathrm{val}}$ denote all samples contained in $\mathcal{E}_{\mathrm{val}}$. Because flow matching introduces stochasticity through the noise action and flow time, let $\xi_j^{(k)}$ denote their realization for validation sample $z_j$ at optimization step $k$. The corresponding stochastic estimate of the clean validation loss is \begin{equation} 
\widehat{\mathcal{L}}_{\mathrm{val}}^{(k)}(\theta) := \frac{1}{|\mathcal{D}_{\mathrm{val}}|} \sum_{z_j\in\mathcal{D}_{\mathrm{val}}} \ell_{\mathrm{FM}} \bigl(\theta;z_j,\xi_j^{(k)}\bigr). 
\label{eq:validation_loss} 
\end{equation} This objective is differentiable and requires no environment interaction. Under this proxy, a useful candidate sample should induce a parameter update that decreases the loss on clean validation behavior.

Consider a candidate sample $z_{i,r}$ encountered at step $k$. Its stochastic gradient and the gradient induced by the clean validation set are 
\begin{equation} \begin{aligned} g_{i,r}^{(k)} &:= \nabla_\theta \ell_{\mathrm{FM}} \bigl(\theta_k;z_{i,r},\xi_{i,r}^{(k)}\bigr),\\ \bar{g}_{\mathrm{val}}^{(k)} &:= \nabla_\theta \widehat{\mathcal{L}}_{\mathrm{val}}^{(k)}(\theta_k) = \frac{1}{|\mathcal{D}_{\mathrm{val}}|} \sum_{z_j\in\mathcal{D}_{\mathrm{val}}} \nabla_\theta \ell_{\mathrm{FM}} \bigl(\theta_k;z_j,\xi_j^{(k)}\bigr). \end{aligned} \label{eq:candidate_and_reference_gradients} 
\end{equation}
Here, $-\bar{g}_{\mathrm{val}}^{(k)}$ is the descent direction preferred by the clean validation set as a whole. To estimate the effect of training on $z_{i,r}$, consider the counterfactual one-step update $\theta_{k,i,r}^{+}=\theta_k-\eta_k g_{i,r}^{(k)}$. Conditioning on the sampled flow-matching randomness and assuming standard local smoothness, a Taylor expansion around $\theta_k$ gives 
\begin{equation}
    \widehat{\mathcal{L}}_{\mathrm{val}}^{(k)}\!\bigl(\theta_{k,i,r}^{+}\bigr)
    = \widehat{\mathcal{L}}_{\mathrm{val}}^{(k)}(\theta_k)
    - \eta_k \left\langle \bar{g}_{\mathrm{val}}^{(k)}, g_{i,r}^{(k)} \right\rangle
    + O\!\left(\eta_k^2 \left\|g_{i,r}^{(k)}\right\|_2^2\right).
    \label{eq:validation_loss_taylor}
\end{equation}
The inner product determines the first-order change in clean
validation loss: positive values indicate alignment with the clean
objective, while negative values indicate conflict
\citep{koh2017understanding,wang2025data}. However, its magnitude also depends on gradient norms, which vary
with the sampled noise action, flow time, and local loss scale.
To measure directional compatibility while reducing sensitivity
to these scales, we therefore normalize both gradients and define the global compatibility score \begin{equation} s_{i,r,\mathrm{global}}^{(k)} := \cos\!\left( g_{i,r}^{(k)}, \bar{g}_{\mathrm{val}}^{(k)} \right) = \frac{ \left\langle g_{i,r}^{(k)}, \bar{g}_{\mathrm{val}}^{(k)} \right\rangle }{ \left\|g_{i,r}^{(k)}\right\|_2 \left\|\bar{g}_{\mathrm{val}}^{(k)}\right\|_2 }. \label{eq:global_gradient_compatibility} \end{equation} This normalization preserves the sign and directional agreement motivated by the first-order analysis while suppressing nuisance variation in gradient scale. This provides a scale-robust, sample-level proxy for supervision utility. However, the global reference averages supervision across potentially unrelated tasks and interaction stages. The next section addresses this limitation by constructing a context-conditioned local validation reference.

\subsection{Task-Semantic and Visual-Context Neighborhood Reference}
\label{sec:local_reference}

The global reference in Equation~\ref{eq:candidate_and_reference_gradients} averages
gradients over the complete validation set. As illustrated in
Fig.~\ref{fig:global_local}, this can be problematic for robot manipulation
data. A single episode typically traverses several interaction phases, such as
reaching, grasping, transporting, and placing, which involve distinct visual
contexts and action modes. Multi-task collections further introduce
heterogeneity in task semantics. Averaging across these modes can mix
irrelevant or conflicting gradients, producing a reference direction that is
poorly matched to the candidate sample being evaluated.

RoboDrop instead constructs a candidate-specific reference through task-semantic and visual-context retrieval. Given candidate sample $z_{i,r}=(o_{i,r},\ell_i,A_{i,r})$, we use T5 instruction embeddings~\citep{raffel2020exploring} to construct a task-semantic validation pool $\mathcal{C}_i$. If all episodes share the same instruction, this step leaves the validation pool unchanged. Within $\mathcal{C}_i$, we compute the visual similarity
$\rho_{i,r,j}=\cos(f_{\mathrm{DINO}}(o_{i,r}),f_{\mathrm{DINO}}(o_j))$ using frozen DINO features and retain the $K_{\mathrm{vis}}$ most similar validation samples:
\begin{equation}
    \mathcal{N}(i,r)
    :=
    \operatorname{TopK}_{K_{\mathrm{vis}}}
    \left(
        \left\{
            z_j\in\mathcal{C}_i:
            \rho_{i,r,j}
        \right\}
    \right).
    \label{eq:visual_context_neighborhood}
\end{equation}
Cross-episode frames that are more similar in DINO~\citep{oquab2023dinov2} feature space exhibit
substantially higher gradient-direction similarity; Appendix~\ref{app:dino_gradient_alignment}
provides the experimental protocol and analysis.

The retrieved validation gradients are aggregated using softmax-normalized
visual similarities. Specifically,
$w_{i,r,j}\propto\exp(\rho_{i,r,j}/\gamma)$ over
$z_j\in\mathcal{N}(i,r)$, where the temperature is set to $\gamma=0.1$ by
default, and
$g_{\mathrm{val},j}^{(k)}
=
\nabla_\theta\ell_{\mathrm{FM}}(\theta_k;z_j,\xi_j^{(k)})$
denotes the gradient of validation sample $z_j$. Replacing the global
validation gradient with this weighted local reference yields
\begin{equation}
    s_{i,r,\mathrm{local}}^{(k)}
    :=
    \cos\!\left(
        g_{i,r}^{(k)},
        \sum_{z_j\in\mathcal{N}(i,r)}
        w_{i,r,j}\,g_{\mathrm{val},j}^{(k)}
    \right),
    \qquad
    w_{i,r,j}
    =
    \frac{
        \exp(\rho_{i,r,j}/\gamma)
    }{
        \displaystyle
        \sum_{z_m\in\mathcal{N}(i,r)}
        \exp(\rho_{i,r,m}/\gamma)
    }.
    \label{eq:local_gradient_compatibility}
\end{equation}
Unlike the global score, this local compatibility evaluates a candidate update
against clean supervision matched to both its task semantics and visual
interaction context.

\subsection{Online In-Run Scoring and Episode-Level Aggregation}
\label{sec:inrun_accumulation}

A compatibility score computed at a fixed checkpoint captures only an
instantaneous update direction, which can change substantially as the policy evolves during post-training. RoboDrop therefore scores candidate samples online along a one-epoch warm-up trajectory rather than evaluating the entire candidate set at a single model state. 
Ideally, the clean validation gradients would be recomputed at every optimization step so that each candidate gradient is compared with a reference at exactly the same model state. Doing so, however, would require an additional gradient computation over the validation set at every step, substantially increasing the cost of the warm-up run. RoboDrop therefore refreshes the validation gradients once every $M$ steps and caches them between refreshes. Before the one-epoch warm-up, candidate samples are shuffled. Let $k(i,r)$ denote the unique optimization step at which sample $z_{i,r}$ enters a minibatch. At step $k(i,r)$, its gradient is compared with the most recently cached local validation reference. Each sample thus receives exactly one compatibility score, which is then
aggregated within its candidate episode:
\begin{equation}
    \bar{s}_{i,r}
    =
    s_{i,r,\mathrm{local}}^{(k(i,r))},
    \qquad
    S_i
    =
    \frac{1}{L_i}
    \sum_{r=1}^{L_i}
    \bar{s}_{i,r}.
    \label{eq:inrun_episode_score}
\end{equation}
Frame-level shuffling makes the optimization step assigned to each sample
independent of its episode identity. Consequently, the samples of a sufficiently long episode probe compatibility at multiple points along the warm-up trajectory, and their average provides a one-pass estimate of that episode's compatibility across evolving model states. The resulting episode-level score distribution is used for data filtering.

\subsection{Optional Automatic Filtering}
\label{sec:gmm_filtering}

The episode scores $\{S_i\}_{i=1}^{N}$ provide a reliability ranking, but
turning this ranking into filtering decisions still requires a
pre-specified removal ratio. To support flexible deployment on real-world
datasets, where the contamination rate is typically unknown and the candidate
set may contain no unreliable episodes, we additionally provide an optional
automatic filtering procedure. Specifically, the procedure uses BIC to select
among a single Gaussian ($K=1$) and two-component GMMs ($K=2$) with either
shared or component-specific variances:
\begin{equation}
    \mathcal{M}^{\star}
    =
    \arg\min_{\mathcal{M}\in
    \{\mathcal{M}_{1},\mathcal{M}_{2,\mathrm{shared}},
    \mathcal{M}_{2,\mathrm{specific}}\}}
    \operatorname{BIC}(\mathcal{M}),
    \qquad
    K^{\star}=K(\mathcal{M}^{\star}).
    \label{eq:gmm_model_selection}
\end{equation}
If $K^{\star}=1$, RoboDrop retains all candidate episodes. If $K^{\star}=2$, we denote the lower-mean component
by $c^{-}=\arg\min_{c\in\{1,2\}}\mu_c$
and interpret it as unreliable supervision, since lower scores indicate weaker
compatibility with the clean validation reference. The retained episode set is
\begin{equation}
    \widehat{\mathcal{E}}_{\mathrm{keep}}
    =
    \begin{cases}
        \mathcal{E}_{\mathrm{cand}},
        & K^{\star}=1,\\[2pt]
        \left\{
            \tau_i\in\mathcal{E}_{\mathrm{cand}}
            \mid p(c^{-}\mid S_i,\mathcal{M}^{\star})\leq q
        \right\},
        & K^{\star}=2,
    \end{cases}
    \label{eq:gmm_filtering}
\end{equation}
where $q$ is a posterior-confidence threshold. It controls the
retention--purity trade-off: a larger value removes only episodes classified as
unreliable with high confidence and retains more data, whereas a
smaller value filters borderline episodes more aggressively to obtain a cleaner
retained set. After filtering, we discard the warm-up policy and post-train a
fresh copy of the pretrained policy on the retained candidate episodes together
with $\mathcal{E}_{\mathrm{val}}$.

\section{Experiments}
\label{sec:experiments}

\subsection{Experimental Setup}
\label{sec:experimental_setup}

\paragraph{Datasets.}
We evaluate RoboDrop on three sources of mixed-quality robot data with
increasing realism. For \emph{LIBERO}~\citep{liu2023libero}, we construct separate action- and
temporal-corruption datasets by corrupting $10\%$ of the episodes in each of
the four task suites. \emph{Robomimic-MH}~\citep{robomimic2021} comprises four tasks with 300 human
teleoperation episodes per task, including 100 collected by proficient
operators. Our \emph{real-world} dataset comprises four tasks with 100 episodes
per task, including 80 collected by proficient operators.
Demonstrations collected by less-skilled operators often contain execution errors and inefficient behavior, making operator proficiency a useful proxy for demonstration quality. We use these provenance labels to evaluate episode-score separability, while assessing the practical utility of curation through downstream policy success. Dataset construction,
quality-label definitions, and validation splits are detailed in
Appendix~\ref{app:datasets}.
\paragraph{Baselines.}
We compare RoboDrop with four data-selection baselines. \emph{Behavior Retrieval} measures similarity to clean demonstrations using
VAE-encoded state--action representations. \emph{DataMIL} estimates data utility through datamodels, using clean
validation loss as a differentiable proxy for policy performance.
\emph{Scizor} identifies suboptimal and redundant
transitions through self-supervised progress estimation and state--action
similarity. \emph{QoQ} scores demonstrations using
gradient-based influence on validation examples at a fixed policy checkpoint without task-semantic or visual-context conditioning. For a fair comparison, \emph{QoQ} uses the final checkpoint from the same one-epoch warm-up run used by RoboDrop.

\paragraph{Training and evaluation.}
Since all methods ultimately assign scalar scores to transitions or episodes, we obtain episode-level scores using each method's native aggregation when available and simple averaging otherwise. We then assess how well these scores separate reliable from unreliable episodes using AUROC and best balanced accuracy. To evaluate the practical effect of curation, we post-train a fresh copy of a pretrained policy on the data retained by each method and report rollout task success. Unless otherwise specified, all experiments use the open-source $\pi_{0.5}$-Base model~\citep{black2025pi05}; methods requiring clean validation references are provided with the same validation episodes. RoboDrop retrieves the $K_{\mathrm{vis}}=10$ nearest validation samples in DINO feature space and refreshes the cached validation reference gradients every $M=200$ optimization steps during warm-up training. The warm-up stage lasts for one epoch, covering every candidate sample once. To reduce the memory cost of scoring, RoboDrop computes gradients only with respect to the action expert parameters and applies CountSketch~\citep{charikar2002finding, schioppa2024efficient, xia2024less} to compress each gradient to 4096 dimensions before computing compatibility scores. Detailed training and evaluation procedures are provided in Appendix~\ref{app:training_details}.

\subsection{Controlled Corruptions on LIBERO}
\label{sec:libero_corruptions}

\begin{figure*}[!t]
    \centering
    \begin{minipage}[t]{0.31\textwidth}
        \vspace{0pt}
        \centering
        \includegraphics[width=\linewidth,height=0.20\textheight,keepaspectratio]{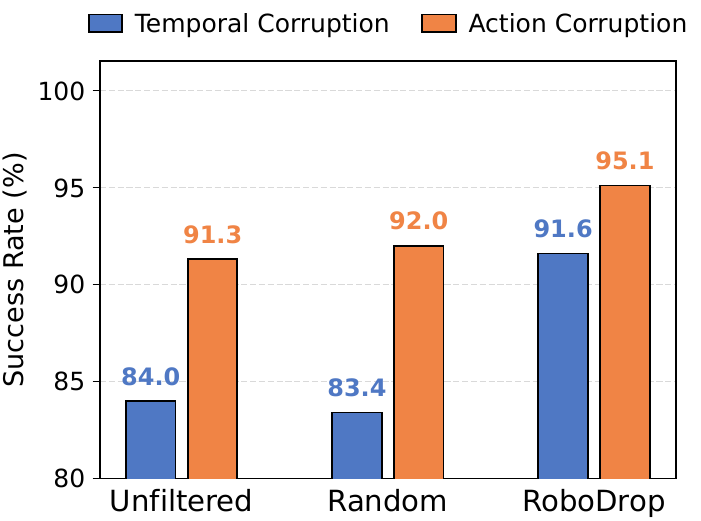}
        \captionof{figure}{LIBERO rollout success rate after data curation.}
        \label{fig:libero_success_rate}
        \label{fig:rollout_success_rates}
    \end{minipage}\hfill
    \begin{minipage}[t]{0.65\textwidth}
        \vspace{0pt}
        \centering
\captionof{table}{Corruption discrimination and downstream
curation performance on LIBERO-10. T and A denote temporal and action corruption. All results are
percentages.}

        \label{tab:libero10_baselines}
        \resizebox{\linewidth}{!}{%
        \begin{tabular}{lccccccccc}
            \toprule
            & \multicolumn{3}{c}{AUROC $\uparrow$}
            & \multicolumn{3}{c}{Best bal. acc. $\uparrow$}
            & \multicolumn{3}{c}{Curated success $\uparrow$} \\
            \cmidrule(lr){2-4}
            \cmidrule(lr){5-7}
            \cmidrule(lr){8-10}
            Method & T & A & Avg. & T & A & Avg. & T & A & Avg. \\
            \midrule
            BR
                & 81.8 & 91.0 & 86.4
                & 78.7 & 86.6 & 82.7
                & 83.5 & 86.0 & 84.8 \\
            DataMIL
                & 68.6 & \textbf{100.0} & 84.3
                & 65.7 & \textbf{100.0} & 82.9
                & 81.0 & \textbf{95.0} & 88.0 \\
            Scizor
                & 47.7 & 47.4 & 47.6
                & 54.9 & 56.0 & 55.5
                & 76.0 & 85.5 & 80.8 \\
            QoQ
                & 61.6 & \textbf{100.0} & 80.8
                & 61.1 & \textbf{100.0} & 80.5
                & 81.0 & \textbf{95.0} & 88.0 \\
            \textbf{RoboDrop}
                & \textbf{96.1} & 98.7 & \textbf{97.4}
                & \textbf{91.6} & 95.3 & \textbf{93.5}
                & \textbf{89.0} & 94.5 & \textbf{91.8} \\
            \bottomrule
        \end{tabular}%
        }
    \end{minipage}
\end{figure*}

We evaluate RoboDrop on all four LIBERO task suites under both temporal and action corruption. In each setting, RoboDrop scores the candidate episodes, removes the lowest-scoring $10\%$, and post-trains a fresh policy on the retained data. We compare its rollout success with post-training on the unfiltered data and with randomly removing the same $10\%$ budget. As shown in Figure~\ref{fig:libero_success_rate}, RoboDrop improves the average success rate from $84.0\%$ (unfiltered) and $83.4\%$ (random removal) to $91.6\%$ under temporal corruption, and from $91.3\%$ and $92.0\%$ to $95.1\%$ under action corruption. These results show that the episode ranking translates directly into more effective post-training across distinct corruption mechanisms. Per-suite AUROC, best balanced accuracy, and rollout-success comparisons are
reported in Appendix~\ref{app:libero_detailed_results}.

\paragraph{Comparison with data-selection baselines.}
Table~\ref{tab:libero10_baselines} compares the episode rankings produced by RoboDrop and the baselines on LIBERO-10. RoboDrop performs consistently across both corruption types. It substantially outperforms all baselines under temporal corruption, achieving an AUROC of $96.1\%$ compared with $81.8\%$ for the strongest baseline. DataMIL and QoQ perfectly separate the controlled action corruption, but deteriorate markedly under temporal misalignment. RoboDrop achieves the strongest average discrimination across the two corruptions and the highest average post-filtering success rate. These results suggest that action-space and fixed-checkpoint signals can be effective for direct action perturbations, whereas detecting observation--action misalignment benefits from a context-conditioned reference evaluated online during training. 

\subsection{Naturally Varying Demonstrations and Backbone Generality}
\label{sec:robomimic_results}

The controlled LIBERO experiments isolate specific corruption mechanisms, but
real demonstration quality is rarely binary or synthetically generated. We
therefore evaluate whether RoboDrop can distinguish naturally occurring
quality variation in Robomimic-MH, where demonstrations are collected from
operators with different proficiency levels. Following the protocol in
Section~\ref{sec:experimental_setup}, we treat the designated proficient
demonstrations as reliable and the remaining demonstrations as suboptimal.
These labels provide an evaluation criterion rather than supervision to the
curation methods.

\begin{figure*}[!t]
    \centering
    \begin{minipage}[t]{0.77\textwidth}
        \vspace{0pt}
        \centering
        \includegraphics[width=\linewidth]{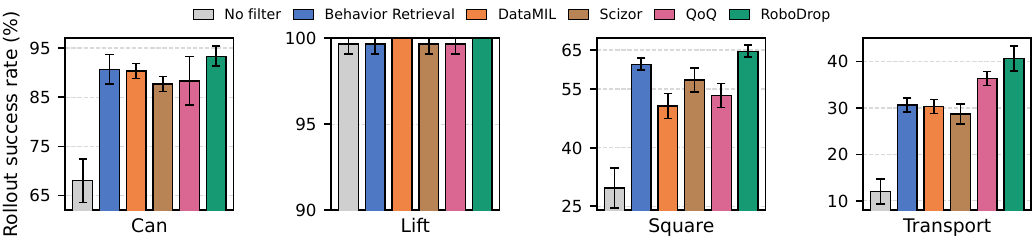}
        \captionof{figure}{Downstream rollout success after curating the
        Robomimic-MH post-training data.}
        \label{fig:robomimic_rollout_success}
    \end{minipage}\hfill
    \begin{minipage}[t]{0.20\textwidth}
        \vspace{0pt}
        \centering
        \includegraphics[width=\linewidth]{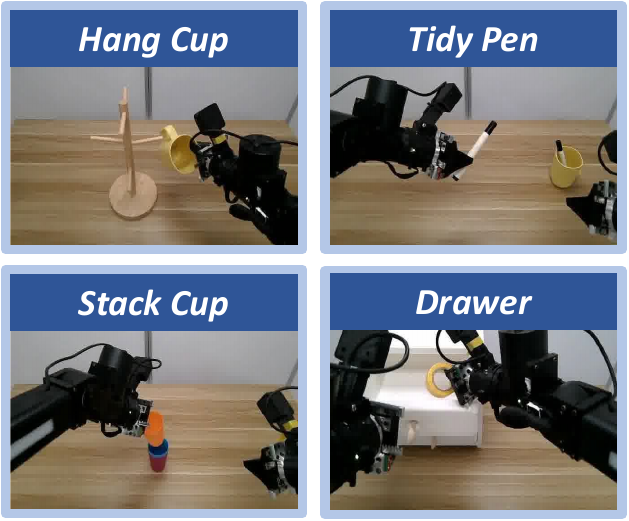}
        \captionof{figure}{The four real-world tasks.}
        \label{fig:realworld_tasks}
    \end{minipage}
\end{figure*}

\paragraph{Comparison with data-selection baselines.}
Table~\ref{tab:robomimic_baselines} summarizes episode-score separability averaged over the four Robomimic-MH tasks. RoboDrop uses $\pi_{0.5}$-Base for scoring and achieves the highest mean AUROC and balanced accuracy, indicating the strongest overall discrimination. We next assess downstream policy performance by discarding the lowest-ranked $50\%$ of demonstrations under each curation method and post-training a fresh X-VLA-Base model on the retained data. As shown in Figure~\ref{fig:robomimic_rollout_success}, RoboDrop achieves the highest success rate on all tasks except Lift, where performance is near saturation. It attains $74.67\%$ mean success across the four tasks, exceeding unfiltered post-training ($52.34\%$) by $22.33$ percentage points. This substantial gain suggests that suboptimal behavior within ultimately successful demonstrations can impair policy learning: completing a task does not ensure that every intermediate observation--action pair provides useful supervision. The results highlight the value of post-training data curation beyond simply retaining successful demonstrations. Detailed per-task results are reported in
Appendix~\ref{app:robomimic_detailed_results}.

\paragraph{Generalization across VLA backbones.}
RoboDrop defines compatibility through gradients of the post-training
objective and is therefore not tied to a particular VLA architecture. We test
this property using $\pi_{0.5}$-Base,
GR00T, and X-VLA~\citep{zheng2026x}.
As shown in Table~\ref{tab:robomimic_backbones}, RoboDrop maintains a mean
AUROC above $85\%$ with each pretrained VLA backbone, with relatively small
variation in mean balanced accuracy.
We further evaluate a weaker X-VLA initialization with a Florence-initialized vision--language backbone and a randomly initialized action expert. RoboDrop retains a mean AUROC of $80.12\%$, suggesting that task-specific training can yield useful compatibility signals without action pretraining. Fully pretrained X-VLA improves mean AUROC and balanced accuracy by $5.23$ and $3.14$ percentage points, indicating that VLA pretraining strengthens these signals but is not essential. Per-task results are provided in
Appendix~\ref{app:robomimic_detailed_results}.

\begin{table*}[!t]
    \centering
    \begin{minipage}[t]{0.49\textwidth}
        \centering
        \captionof{table}{Suboptimal-episodes discrimination on Robomimic-MH (\%).}
        \label{tab:robomimic_baselines}
        \footnotesize
        \setlength{\tabcolsep}{4pt}
        \begin{tabular}{lcc}
            \toprule
            Method & AUROC $\uparrow$ & Best bal. acc. $\uparrow$ \\
            \midrule
            Behavior Retrieval & 80.99 & 75.02 \\
            DataMIL & 76.46 & 72.00 \\
            Scizor & 80.52 & 77.24 \\
            QoQ & 83.67 & 76.20 \\
            \textbf{RoboDrop} & \textbf{88.33} & \textbf{81.83} \\
            \bottomrule
        \end{tabular}
    \end{minipage}\hfill
    \begin{minipage}[t]{0.49\textwidth}
        \centering
        \captionof{table}{Suboptimal-episodes discrimination for RoboDrop across VLA backbones on
        Robomimic-MH (\%).}
        \label{tab:robomimic_backbones}
        \footnotesize
        \setlength{\tabcolsep}{4pt}
        \begin{tabular}{lcc}
            \toprule
            Backbone & AUROC $\uparrow$ & Best bal. acc. $\uparrow$ \\
            \midrule
            $\pi_{0.5}$-Base & \textbf{88.33} & \textbf{81.83} \\
            GR00T & 85.87 & 81.21 \\
            X-VLA & 85.35 & 78.46 \\
            X-VLA (Florence) & 80.12 & 75.32 \\
            \bottomrule
        \end{tabular}
    \end{minipage}
\end{table*}

\FloatBarrier
\subsection{Real-World Data Curation}
\label{sec:real_world_results}

\begin{wraptable}[8]{r}{0.50\textwidth}
    \vspace{-9pt}
    \centering
    \footnotesize
    \captionsetup{skip=1pt}
    \setlength{\tabcolsep}{5.5pt}
    \renewcommand{\arraystretch}{0.94}
    \caption{Average non-expert identification on real-robot tasks (\%).}
    \label{tab:realworld_detection}
    \begin{tabular}{lcc}
        \toprule
        Method & AUROC $\uparrow$ & Best bacc $\uparrow$ \\
        \midrule
        Behavior Retrieval  & 94.77 & 90.89 \\
        DataMIL             & 86.31 & 85.09 \\
        Scizor              & 79.49 & 80.00 \\
        QoQ                 & 83.31 & 82.14 \\
        \textbf{RoboDrop}   & \textbf{98.38} & \textbf{95.63} \\
        \bottomrule
    \end{tabular}
    \vspace{-7pt}
\end{wraptable}

We finally evaluate RoboDrop on demonstrations collected through real-robot
teleoperation, where collection errors arise naturally. The four tasks are illustrated in Figure~\ref{fig:realworld_tasks}. Detailed task descriptions and the complete dataset and validation split are
provided in Appendix~\ref{app:realworld_dataset}.

\vspace{-3pt}
\paragraph{Comparison with data-selection baselines.}
We first examine whether RoboDrop remains sensitive to demonstration quality when errors arise naturally during real-robot collection. As shown in Table~\ref{tab:realworld_detection}, RoboDrop achieves the highest average AUROC and best balanced accuracy across the four tasks. The gains over the best baseline Behavior Retrieval indicate that local gradient compatibility goes beyond representation-level similarity, better capturing whether a demonstration induces an update consistent with clean, task-relevant supervision.  We next examine whether the stronger episode-level discrimination translates into improved downstream policies. To compare episode-ranking quality independently of budget selection, we apply a common 20\% removal budget to all methods and remove the lowest-scoring episodes under each ranking. We post-train a fresh $\pi_{0.5}$-Base policy on each curated dataset and evaluate $20$ rollouts per task. As shown in Table~\ref{tab:realworld_success_comparison}, RoboDrop improves average success from $35.00\%$ without filtering to $63.75\%$, outperforming all baselines. These results show that the quality distinctions captured by local gradient compatibility translate into more effective post-training data selection.

\vspace{-7pt}
\paragraph{Budget selection in practice.} The fixed-budget comparison above evaluates which episodes are selected under
a common removal budget, but assumes prior knowledge of how much data to
remove. As a practical implementation when this information is unavailable, we fit one- and two-component Gaussian mixture models to the RoboDrop episode scores, select between them using BIC, and remove episodes assigned to the lower-score component with posterior confidence above $q=0.8$. Without access to the contamination rate, this procedure achieves a competitive average success rate of $67.50\%$ relative to manually specified fixed budgets. Additional diagnostics across contamination rates from $10\%$ to $40\%$ are reported in Appendix~\ref{app:realworld_filtering_details}.

\begin{table*}[!t]
    \centering
    \footnotesize
    \setlength{\tabcolsep}{2.0pt}
    \renewcommand{\arraystretch}{1.02}
    \captionsetup{skip=2pt}
    \caption{Per-task real-world rollout success after curation (successful
    rollouts / 20). }
    \label{tab:realworld_success_comparison}
    \begin{tabular}{lccccccccc}
        \toprule
        & & \multicolumn{5}{c}{Fixed $20\%$ removal}
        & \multicolumn{3}{c}{RoboDrop variants} \\
        \cmidrule(lr){3-7}
        \cmidrule(lr){8-10}
        Task & Unfiltered & BR & DataMIL & Scizor & QoQ & RoboDrop
        & Fixed $10\%$ & Fixed $30\%$ & Automatic \\
        \midrule
        \textsc{Hang Cup}  & 2/20  & 8/20  & 5/20  & 4/20  & 5/20  & 9/20  & 6/20  & 8/20  & 9/20 \\
        \textsc{Tidy Pen}  & 7/20  & 12/20 & 8/20  & 8/20  & 9/20  & 12/20 & 9/20  & 12/20 & 14/20 \\
        \textsc{Stack Cup} & 10/20 & 13/20 & 11/20 & 14/20 & 12/20 & 14/20 & 14/20 & 13/20 & 14/20 \\
        \textsc{Drawer}    & 9/20  & 14/20 & 15/20 & 10/20 & 11/20 & 16/20 & 11/20 & 16/20 & 17/20 \\
        \midrule
        Average SR (\%) & 35.0 & 58.75 & 48.75 & 45.0 & 46.25 & 63.75
        & 50.0 & 61.25 & \textbf{67.5} \\
        \bottomrule
    \end{tabular}
\end{table*}

\vspace{-4pt}
\subsection{Ablation Studies}
\label{sec:ablations}

\begin{table}[!t]
    \centering
    \footnotesize
    \captionsetup{skip=5pt}
    \setlength{\tabcolsep}{7pt}
    \renewcommand{\arraystretch}{1.18}
    \caption{Robomimic-MH \textsc{Can} ablations over three runs with different
    clean sets and training seeds (mean $\pm$ std, \%).}
    \label{tab:robodrop_ablations}
    \begin{tabular}{@{}lcc@{}}
        \toprule
        Variant & AUROC $\uparrow$ & Best bacc $\uparrow$ \\
        \midrule
        Global gradient reference & $79.64 \pm 0.36$ & $73.21 \pm 0.66$ \\
        Fixed checkpoint           & $83.50 \pm 0.39$ & $75.81 \pm 0.43$ \\
        DINO similarity only       & $75.86 \pm 1.00$ & $69.23 \pm 1.29$ \\
        DINO + Action similarity   & $79.73 \pm 0.26$ & $73.79 \pm 0.74$ \\
        Dot product score          & $81.78 \pm 2.44$ & $74.60 \pm 2.24$ \\
        \textbf{RoboDrop}          & $\mathbf{87.67 \pm 0.87}$ & $\mathbf{80.82 \pm 1.45}$ \\
        \bottomrule
    \end{tabular}
\end{table}

We ablate RoboDrop on the Robomimic-MH \textsc{Can} task using $\pi_{0.5}$-Base over three runs with different clean validation sets and training seeds. RoboDrop achieves $87.67\pm0.87$ AUROC and $80.82\pm1.45$ balanced accuracy, indicating stable performance under changes in both reference-set composition and training stochasticity. As shown in Table~\ref{tab:robodrop_ablations}, replacing the local gradient reference with a gradient averaged over the full validation set lowers AUROC and balanced accuracy by $8.03$ and $7.61$ points. This suggests that global averaging obscures state-relevant supervision, supporting a local reference for evaluating update compatibility. Fixed-checkpoint scoring evaluates all candidates at the checkpoint obtained after the one-epoch warm-up run and achieves $83.50$ AUROC and $75.81$ balanced accuracy, supporting online scoring along the evolving warm-up trajectory rather than evaluation at its final model state.

We further compare two non-gradient alternatives. \emph{DINO similarity only} scores each candidate using the softmax-weighted visual similarities of its $10$ nearest validation samples, while \emph{DINO + Action similarity} additionally incorporates action similarity over the subsequent $10$ steps.
Adding action information improves AUROC from $75.86$ to $79.73$, but remains $7.94$ points below RoboDrop. This suggests that visual and action similarity provide useful but incomplete quality signals; gradient compatibility instead relates observations and actions through the policy loss to assess their induced update directions. Finally, replacing cosine similarity with the unnormalized gradient dot product lowers AUROC and balanced accuracy by $5.89$ and $6.22$ points. Cosine also reduces the AUROC standard deviation from $2.44$ to $0.87$, supporting normalization of overall gradient scales for both discrimination and stability. Additional sensitivity analyses are reported in Appendix~\ref{app:additional_ablations}.

\subsection{Robustness to reference composition}
\begin{figure}[!htbp]
    \centering
    \captionsetup[subfigure]{skip=1pt}
    \begin{subfigure}[t]{0.24\textwidth}
        \centering
        \includegraphics[width=\linewidth]{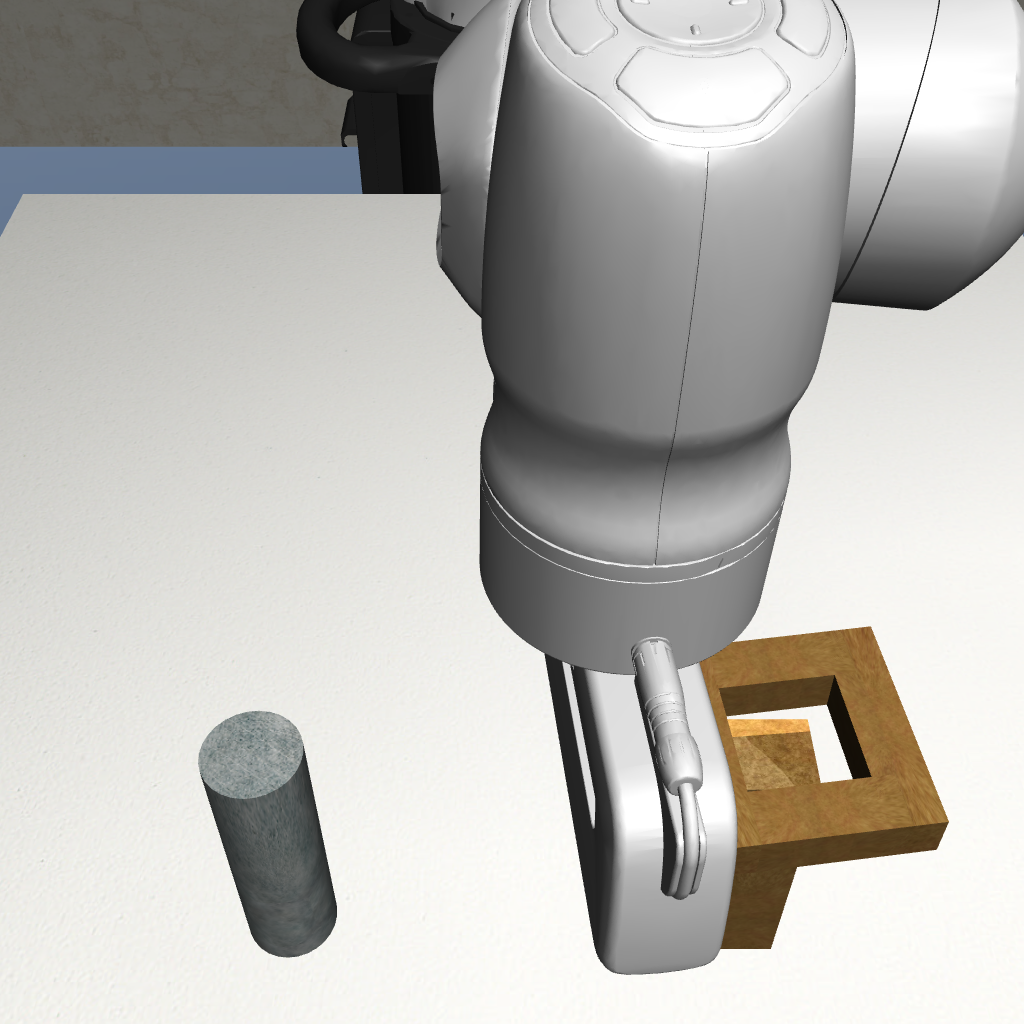}
        \caption{Left}
    \end{subfigure}\hspace{0.03\textwidth}
    \begin{subfigure}[t]{0.24\textwidth}
        \centering
        \includegraphics[width=\linewidth]{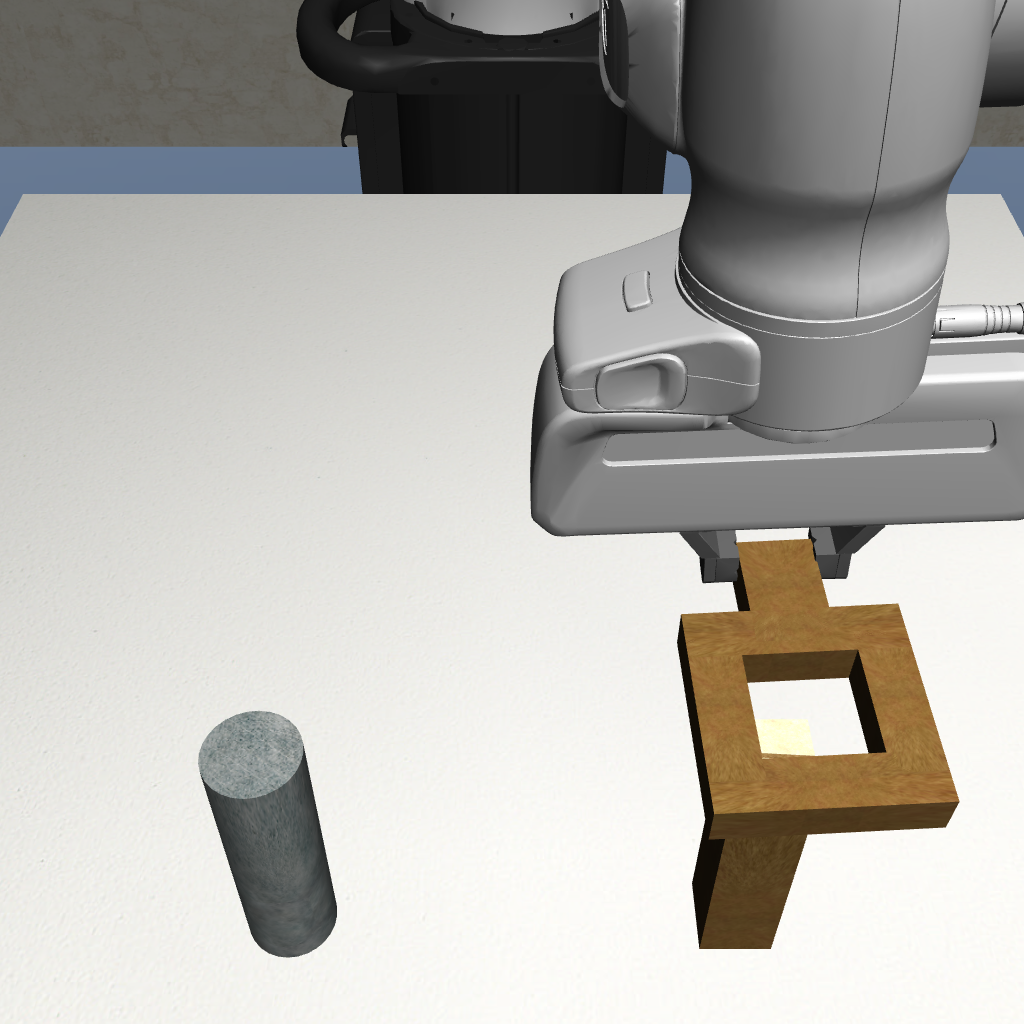}
        \caption{Front}
    \end{subfigure}\hspace{0.03\textwidth}
    \begin{subfigure}[t]{0.24\textwidth}
        \centering
        \includegraphics[width=\linewidth]{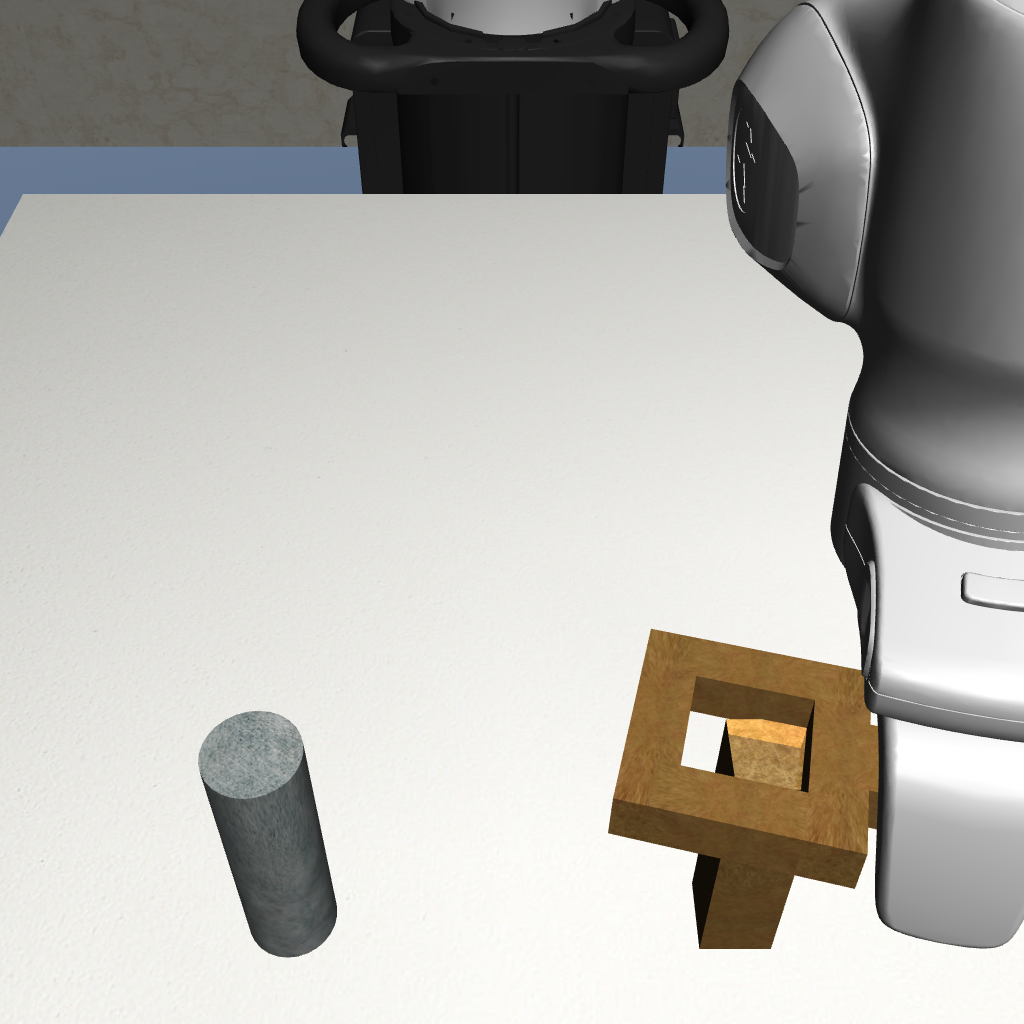}
        \caption{Right}
    \end{subfigure}
    \caption{Three valid manipulation modes on Robomimic-MH \textsc{Square}.}
    \label{fig:square_modes}
\end{figure}

To evaluate RoboDrop's robustness to reference selection, we examine whether
the action modes represented in the reference set affect data selection and
downstream policy success on Robomimic-MH \textsc{Square}. We group the
demonstrations into three manipulation modes, Left, Front, and Right, as
illustrated in Figure~\ref{fig:square_modes}. These modes contain 53, 125, and
100 episodes, respectively. The remaining 22 episodes involve unsuccessful
insertion attempts or atypical insertion configurations.

We construct three reference sets, each containing ten proficient
demonstrations. The Left:Front:Right composition is $3:1:6$ for
\emph{Random}, $0:0:10$ for \emph{Right-only}, and $2:4:4$ for
\emph{Stratified}. For each reference set, RoboDrop
scores the candidate demonstrations, removes the lowest-scoring $50\%$, and
post-trains a policy on the curated data. The policy obtained under each
sampling strategy is evaluated over three rounds of 100 rollouts.
Table~\ref{tab:reference_composition} reports the retained proficient
candidates for each mode, the fraction of non-proficient candidates removed,
and downstream rollout success.

As shown in Table~\ref{tab:reference_composition}, the number of retained
proficient demonstrations in each mode varies only slightly despite large
differences in reference composition. Even under the strongly biased
Right-only strategy, RoboDrop retains 35/37 proficient Front candidates and
10/17 proficient Left candidates, preserving $45/54=83.33\%$ of proficient
demonstrations across the two unrepresented modes. These results indicate
that RoboDrop can retain valid behaviors beyond those explicitly represented
in the reference set. However, the lower Left-mode retention rate also shows
that extreme reference bias can affect individual modes. Downstream rollout
success remains similar across the three configurations, suggesting that the
modest differences in retained mode counts have a limited effect on policy
performance in this setting. Together, these results support RoboDrop's
robustness to reference composition.

\begin{table}[!htbp]
    \centering
    \footnotesize
    \setlength{\tabcolsep}{5pt}
    \caption{Reference-composition experiment on Robomimic-MH \textsc{Square}.
    Retention entries are retained / candidate proficient episodes, excluding
    reference episodes.}
    \label{tab:reference_composition}
    \begin{tabular}{lccccc}
        \toprule
        \multirow{2}{*}{Reference set}
        & \multicolumn{3}{c}{Proficient retention}
        & \multirow{2}{*}{Non-proficient removal (\%) $\uparrow$}
        & \multirow{2}{*}{Rollout success (\%) $\uparrow$} \\
        \cmidrule(lr){2-4}
        & Left & Front & Right & & \\
        \midrule
        Random & 12/14 & 34/36 & 34/40 & 67.50 & $62.33 \pm 2.05$ \\
        Right-only & 10/17 & 35/37 & 34/36 & 67.00 & $61.33 \pm 1.25$ \\
        Stratified & 12/15 & 32/33 & 34/42 & 66.50 & $63.67 \pm 1.70$ \\
        \bottomrule
    \end{tabular}
\end{table}

\FloatBarrier
\vspace{-6pt}
\section{Conclusion}
\label{sec:conclusion}
RoboDrop curates VLA post-training data by auditing supervision through context-conditioned local gradient compatibility evaluated along the warm-up trajectory. Across controlled corruptions, naturally varying demonstrations and real-robot collection errors, RoboDrop consistently discriminates data quality and improves downstream policy success, establishing training-dynamics-aware supervision auditing as a practical tool for VLA data curation. A remaining limitation is its reliance on a clean reference set, whose collection or verification may still require expert effort. Substantial reference bias may also affect the filtered data and the learned policy's action-mode preferences. Reducing this dependence is an important direction for future work.

\clearpage
\bibliography{iclr2027_conference}
\bibliographystyle{iclr2027_conference}

\clearpage
\appendix
\raggedbottom
\captionsetup{skip=6pt}
\setlength{\textfloatsep}{10pt plus 2pt minus 2pt}
\setlength{\floatsep}{8pt plus 2pt minus 2pt}
\setlength{\intextsep}{8pt plus 2pt minus 2pt}
\section{Datasets}
\label{app:datasets}

\subsection{LIBERO}
\label{app:libero_dataset}
LIBERO is a benchmark for lifelong robot manipulation comprising four suites with ten tasks each. LIBERO-Spatial evaluates sensitivity to spatial relationships between objects, LIBERO-Object emphasizes generalization across different object categories, LIBERO-Goal tests the execution of diverse language-specified goals in related scenes, and LIBERO-Long focuses on long-horizon tasks requiring multiple sequential manipulation skills, with each suite containing ten tasks. We construct the action- and
temporal-corruption datasets separately by corrupting $10\%$ of the episodes
for every task. Action corruption adds zero-mean Gaussian noise with standard
deviation $0.25\sigma_a$ to the demonstrated actions, where $\sigma_a$ is the
empirical action standard deviation. Temporal corruption shifts the action
sequence by two seconds relative to the observations. For validation, we
reserve two uncorrupted episodes per task, giving 20 clean validation episodes
for each suite.

\subsection{Robomimic-MH}
\label{app:robomimic_dataset}
Robomimic-MH contains 300 human teleoperation episodes for each of the four
tasks. We designate 100 proficient episodes per task as reliable and treat the
remaining 200 as naturally suboptimal. The proficient episodes correspond to
the \texttt{better} split for most tasks; for \textsc{Transport}, we combine
the \texttt{better} and \texttt{okay\_better} splits to obtain 100 episodes.
We reserve ten proficient episodes per task as clean validation episodes,
leaving 290 candidate episodes for scoring. These provenance labels are used
only for evaluation and are not supplied to the curation methods.

\subsection{Real-World Dataset}
\label{app:realworld_dataset}
We collected 100 teleoperated episodes for each of four real-world tasks using ARX-series robot arms. For each task, proficient operators collected 80 episodes and non-expert operators collected 20. We reserved ten proficient episodes as the reference set and manually checked each episode for successful task completion, adherence to the task's standard operating procedure (SOP), and observable execution issues such as unintended collisions, grasp failures, and redundant motions. RoboDrop requires only that these episodes provide reliable supervision for the intended task, regardless of whether they are optimal for downstream policy learning. In practice, reference demonstrations can be obtained during the pilot collection used to establish the task SOP, allowing manual inspection to focus on a small subset. The remaining 90 episodes per task comprise 70 proficient and 20 non-expert demonstrations. The non-expert subset includes suboptimal trajectories with grasp failures, inefficient motions, and other execution errors. For these candidate episodes, operator provenance is used only for evaluation and is withheld from RoboDrop and all baselines during scoring.

\begin{figure}[!htbp]
    \centering
    \includegraphics[width=0.96\textwidth]{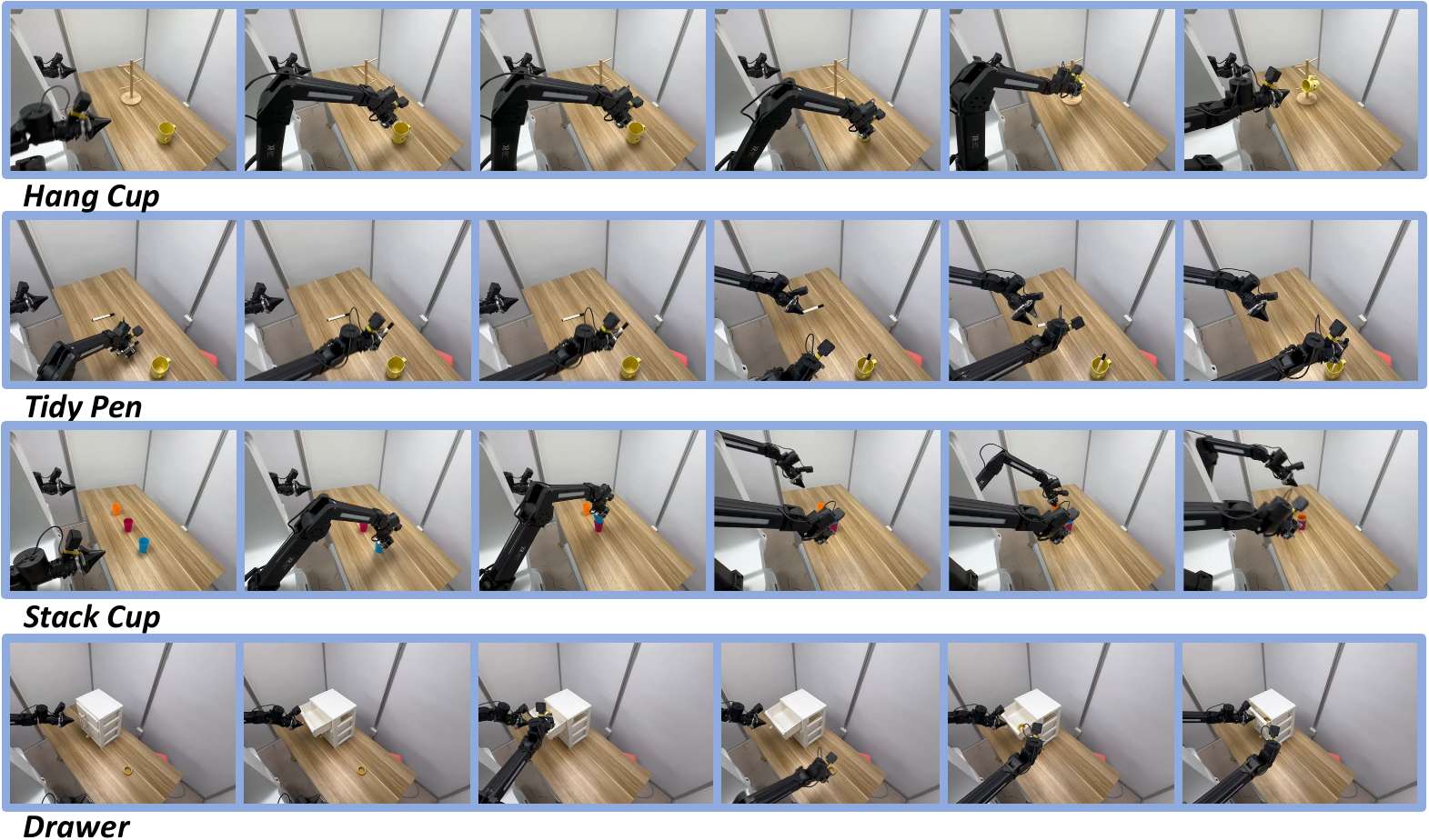}
    \caption{Detailed execution sequences for the four real-world manipulation
    tasks. Each row shows the principal stages of one task.}
    \label{fig:realworld_task_details}
\end{figure}

Figure~\ref{fig:realworld_task_details} shows representative execution
sequences for all four tasks. In \textsc{Hang Cup}, one arm grasps the cup and
places the opening of its handle over a branch of the cup rack. In
\textsc{Tidy Pen}, the right arm first picks up the marker on the right and
places it in the pen holder; the left arm then picks up the marker on the left,
hands it to the right arm, and the right arm places it in the holder. In
\textsc{Stack Cup}, the right arm stacks the rightmost cup onto the center cup,
after which the left arm places the leftmost cup on top. In \textsc{Drawer},
the left arm opens the drawer, the right arm picks up the roll of tape and
places it inside, and the left arm closes the drawer.

\section{Detailed Results}
\label{app:detailed_results}
\subsection{Detailed LIBERO Results}
\label{app:libero_detailed_results}

Table~\ref{tab:libero_all_suites} provides the complete breakdown for the four
LIBERO task suites and both corruption types. RoboDrop maintains strong score
separability across all eight suite--corruption settings, with an average AUROC
of $96.20\%$ and best balanced accuracy of $93.13\%$. Its ranking also yields
consistent downstream gains: removing the lowest-scoring $10\%$ of episodes
raises average rollout success from $87.63\%$ with unfiltered post-training and
$87.69\%$ with random removal to $93.38\%$. The improvement is larger under
temporal corruption ($+7.63$ percentage points over unfiltered data) than under
action corruption ($+3.88$ points), consistent with temporal misalignment being
particularly harmful when left unfiltered. RoboDrop also recovers most of the
$95.69\%$ success achieved by the oracle that removes exactly the corrupted
episodes.

\begin{table}[H]
    \centering
    \small
    \setlength{\tabcolsep}{4.5pt}
    \renewcommand{\arraystretch}{1.0}
    \caption{Per-suite controlled-corruption results on LIBERO (\%).}
    \label{tab:libero_all_suites}
    \begin{tabular}{llccrrrr}
        \toprule
        & & \multicolumn{2}{c}{Score separability}
        & \multicolumn{4}{c}{Rollout success} \\
        \cmidrule(lr){3-4}\cmidrule(lr){5-8}
        Suite & Corruption
        & AUROC
        & Best bal. acc.
        & Unfiltered & Random & \textbf{RoboDrop} & Oracle \\
        \midrule
        LIBERO-Object
        & Temporal & 98.27 & 93.84 & 87.5 & 82.5 & 95.0 & 96.0 \\
        & Action   & 98.78 & 96.09 & 97.5 & 95.5 & 97.5 & 97.5 \\
        \midrule
        LIBERO-Goal
        & Temporal & 93.96 & 90.33 & 79.5 & 82.5 & 89.0 & 94.0 \\
        & Action   & 87.66 & 89.71 & 87.5 & 91.5 & 95.0 & 95.5 \\
        \midrule
        LIBERO-Spatial
        & Temporal & 98.46 & 93.96 & 85.5 & 89.0 & 93.5 & 97.5 \\
        & Action   & 97.68 & 94.11 & 92.5 & 93.0 & 93.5 & 97.0 \\
        \midrule
        LIBERO-10
        & Temporal & 96.11 & 91.62 & 83.5 & 79.5 & 89.0 & 93.5 \\
        & Action   & 98.70 & 95.34 & 87.5 & 88.0 & 94.5 & 94.5 \\
        \midrule
        \textbf{Average}
        & Temporal & 96.70 & 92.44 & 84.00 & 83.38 & \textbf{91.63} & 95.25 \\
        & Action   & 95.71 & 93.81 & 91.25 & 92.00 & \textbf{95.13} & 96.13 \\
        \bottomrule
    \end{tabular}
\end{table}

\FloatBarrier
\subsection{Detailed Robomimic Results}
\label{app:robomimic_detailed_results}

Tables~\ref{tab:robomimic_baselines_detailed} and
\ref{tab:robomimic_backbones_detailed} provide the per-task results underlying
the four-task averages in Tables~\ref{tab:robomimic_baselines} and
\ref{tab:robomimic_backbones}. For data selection, RoboDrop achieves the best
AUROC on \textsc{Can} and \textsc{Square}, remains close to the strongest
method on \textsc{Lift} and \textsc{Transport}, and obtains the best balanced
accuracy on three of the four tasks. In contrast, the strongest baselines vary
by task: Scizor performs best on \textsc{Lift}, while QoQ attains the highest
AUROC on \textsc{Transport}. This variation explains why RoboDrop's consistent
task-level performance yields the strongest overall averages.

Across backbones, $\pi_{0.5}$-Base gives the strongest results on
\textsc{Can} and \textsc{Square} and the highest mean performance. GR00T is
competitive on \textsc{Lift}, while X-VLA performs best on \textsc{Transport},
showing that the compatibility signal transfers across architectures despite
task-specific variation. The Florence-initialized X-VLA remains above chance
on all tasks but is consistently weaker on average, supporting the benefit of
a pretrained action model while confirming that RoboDrop does not require a
particular backbone.

Table~\ref{tab:robomimic_rollout_detailed} reports the per-task rollout results
underlying Figure~\ref{fig:robomimic_rollout_success}. Methods follow the same
order as the figure legend, and the average is the arithmetic mean of the four
task means.

\begin{table}[!htbp]
    \centering
    \footnotesize
    \setlength{\tabcolsep}{4.2pt}
    \renewcommand{\arraystretch}{1.02}
    \caption{Per-task downstream rollout success on Robomimic-MH
    (mean $\pm$ std, \%).}
    \label{tab:robomimic_rollout_detailed}
    \begin{tabular}{lccccc}
        \toprule
        Method
        & \textsc{Can}
        & \textsc{Lift}
        & \textsc{Square}
        & \textsc{Transport}
        & Average \\
        \midrule
        No filter
        & $68.00 \pm 4.42$
        & $99.67 \pm 0.58$
        & $29.67 \pm 5.13$
        & $12.00 \pm 2.65$
        & 52.34 \\
        Behavior Retrieval
        & $90.67 \pm 3.04$
        & $99.67 \pm 0.58$
        & $61.33 \pm 1.53$
        & $30.67 \pm 1.53$
        & 70.59 \\
        DataMIL
        & $90.33 \pm 1.53$
        & $100.00 \pm 0.00$
        & $50.67 \pm 3.21$
        & $30.33 \pm 1.50$
        & 67.83 \\
        Scizor
        & $87.67 \pm 1.53$
        & $99.67 \pm 0.58$
        & $57.33 \pm 3.06$
        & $28.67 \pm 2.14$
        & 68.34 \\
        QoQ
        & $88.33 \pm 4.93$
        & $99.67 \pm 0.58$
        & $53.33 \pm 3.06$
        & $36.33 \pm 1.53$
        & 69.42 \\
        \textbf{RoboDrop}
        & $\mathbf{93.33 \pm 2.03}$
        & $\mathbf{100.00 \pm 0.00}$
        & $\mathbf{64.67 \pm 1.53}$
        & $\mathbf{40.67 \pm 2.68}$
        & \textbf{74.67} \\
        \bottomrule
    \end{tabular}
\end{table}

\begin{table}[!htbp]
    \centering
    \footnotesize
    \setlength{\tabcolsep}{5pt}
    \caption{Per-task identification of naturally suboptimal Robomimic-MH
    demonstrations. Entries are AUROC / best balanced accuracy (\%).}
    \label{tab:robomimic_baselines_detailed}
    \begin{tabular}{lccccc}
        \toprule
        Method
        & \textsc{Can}
        & \textsc{Lift}
        & \textsc{Square}
        & \textsc{Transport}
        & Average \\
        \midrule
        Behavior Retrieval
        & 85.59 / 75.89
        & 88.54 / 79.58
        & 82.61 / 77.94
        & 67.22 / 66.67
        & 80.99 / 75.02 \\

        DataMIL
        & 78.86 / 74.89
        & 79.04 / 71.39
        & 74.84 / 71.89
        & 73.11 / 69.83
        & 76.46 / 72.00 \\

        Scizor
        & 79.29 / 76.33
        & \textbf{90.51 / 89.72}
        & 87.63 / 81.06
        & 64.63 / 61.86
        & 80.52 / 77.24 \\

        QoQ
        & 83.41 / 75.39
        & 82.78 / 74.83
        & 80.17 / 74.00
        & \textbf{88.32} / 80.58
        & 83.67 / 76.20 \\

        \textbf{RoboDrop}
        & \textbf{88.88 / 82.75}
        & 87.23 / 79.44
        & \textbf{89.07 / 83.14}
        & 88.15 / \textbf{81.97}
        & \textbf{88.33 / 81.83} \\
        \bottomrule
    \end{tabular}
\end{table}

\begin{table}[!htbp]
    \centering
    \footnotesize
    \setlength{\tabcolsep}{5pt}
    \caption{Per-task RoboDrop results across VLA backbones on Robomimic-MH.
    Entries are AUROC / best balanced accuracy (\%).}
    \label{tab:robomimic_backbones_detailed}
    \begin{tabular}{lccccc}
        \toprule
        Backbone
        & \textsc{Can}
        & \textsc{Lift}
        & \textsc{Square}
        & \textsc{Transport}
        & Average \\
        \midrule
        $\pi_{0.5}$-Base
        & \textbf{88.88 / 82.75}
        & \textbf{87.23} / 79.44
        & \textbf{89.07 / 83.14}
        & 88.15 / 81.97
        & \textbf{88.33 / 81.83} \\

        GR00T
        & 87.87 / 81.00
        & 86.78 / \textbf{82.64}
        & 83.10 / 79.97
        & 85.72 / 81.22
        & 85.87 / 81.21 \\

        X-VLA
        & 88.46 / 80.39
        & 74.51 / 70.11
        & 88.74 / 79.39
        & \textbf{89.68 / 83.94}
        & 85.35 / 78.46 \\

        X-VLA (Florence init.)
        & 77.84 / 72.64
        & 74.15 / 68.22
        & 81.73 / 77.14
        & 86.74 / 83.28
        & 80.12 / 75.32 \\
        \bottomrule
    \end{tabular}
\end{table}

\FloatBarrier
\subsection{Detailed Real-World Filtering Results}
\label{app:realworld_filtering_details}

Table~\ref{tab:realworld_detection_detailed} gives the per-task results
underlying the averages in Table~\ref{tab:realworld_detection}. RoboDrop obtains
the strongest AUROC and best balanced accuracy on every task. In contrast, the
relative strength of the baselines varies by task, indicating that no single
trajectory-level heuristic consistently captures non-expert behavior.

\begin{table}[!htbp]
    \centering
    \footnotesize
    \setlength{\tabcolsep}{3.2pt}
    \renewcommand{\arraystretch}{1.0}
    \caption{Per-task real-robot non-expert identification results. Entries are
    AUROC / best balanced accuracy (\%).}
    \label{tab:realworld_detection_detailed}
    \begin{tabular}{lccccc}
        \toprule
        Method & \textsc{Hang Cup} & \textsc{Tidy Pen} & \textsc{Stack Cup}
        & \textsc{Drawer} & Average \\
        \midrule
        Behavior Retrieval
        & 99.57 / 98.57 & 93.00 / 85.00 & 95.79 / 92.14
        & 90.71 / 87.86 & 94.77 / 90.89 \\
        DataMIL
        & 84.50 / 82.14 & 80.79 / 85.36 & 87.86 / 88.57
        & 92.07 / 84.29 & 86.31 / 85.09 \\
        Scizor
        & 62.50 / 71.43 & 81.32 / 76.79 & 98.32 / 96.79
        & 75.82 / 75.00 & 79.49 / 80.00 \\
        QoQ
        & 81.50 / 85.36 & 81.43 / 81.07 & 91.04 / 86.07
        & 79.25 / 76.07 & 83.31 / 82.14 \\
        \textbf{RoboDrop}
        & \textbf{100.00 / 100.00} & \textbf{97.57 / 92.86}
        & \textbf{99.79 / 97.86} & \textbf{96.14 / 91.79}
        & \textbf{98.38 / 95.63} \\
        \bottomrule
    \end{tabular}
\end{table}

\paragraph{Adapting to unknown contamination rates.}
A discriminative ranking alone does not determine how many episodes should be
removed. We therefore test whether RoboDrop's automatic procedure adapts when
the prevalence of non-expert data changes. For each task, we construct subsets
at contamination rates ranging from $10\%$ to $40\%$ using multiple random
samplings. We use the same posterior confidence threshold $q=0.8$ in all
settings without tuning it to the contamination rate. As shown in
Table~\ref{tab:realworld_contamination_sensitivity}, the BIC-selected procedure
achieves macro-average F1 scores from $85.6\%$ to $93.0\%$ while maintaining
precision above $84.0\%$ and recall above $88.2\%$. Thus, its filtering
decisions remain effective as the unknown contamination rate varies.

\begin{table}[H]
    \centering
    \footnotesize
    \setlength{\tabcolsep}{7pt}
    \renewcommand{\arraystretch}{1.15}
    \caption{Automatic filtering under varying non-expert rates (\%).}
    \label{tab:realworld_contamination_sensitivity}
    \begin{tabular}{lccc}
        \toprule
        Non-expert rate & Precision & Recall & F1 \\
        \midrule
        10\% & $84.0 \pm 7.6$ & $92.9 \pm 5.6$ & $85.6 \pm 6.8$ \\
        20\% & $93.4 \pm 1.2$ & $88.2 \pm 1.0$ & $90.5 \pm 1.2$ \\
        30\% & $92.3 \pm 1.4$ & $90.6 \pm 0.6$ & $91.3 \pm 0.6$ \\
        40\% & $95.3 \pm 4.1$ & $91.2 \pm 1.5$ & $93.0 \pm 1.3$ \\
        \bottomrule
    \end{tabular}
\end{table}

Table~\ref{tab:realworld_filtering} isolates the accuracy of the automatic
filtering procedure. A fixed $10\%$ budget achieves perfect precision but only
$45.0\%$ recall on every task, whereas a fixed $30\%$ budget improves recall at
the cost of substantially lower precision. Without access to the contamination
rate, RoboDrop automatically removes 17--21 episodes per task and achieves mean
precision, recall, and F1 scores of $93.78\%$, $90.00\%$, and $91.70\%$,
respectively, maintaining strong filtering accuracy across all four tasks
without a fixed removal budget.

\begin{table}[H]
    \centering
    \caption{Per-task filtering accuracy on 90 non-validation real-robot
    episodes. Fixed $p\%$ is computed over these 90 candidates (e.g.,
    $20\%=18$ removed); metrics are percentages.}
    \label{tab:realworld_filtering}
    \footnotesize
    \setlength{\tabcolsep}{2.2pt}
    \renewcommand{\arraystretch}{1.15}
    \begin{minipage}[t]{0.49\textwidth}
        \centering
        \begin{tabular}{llcccc}
            \toprule
            Task & Rule & Rem. & Prec. & Rec. & F1 \\
            \midrule
            \textsc{Stack Cup} & Auto. & 19 & 100.0 & 95.0 & \textbf{97.4} \\
              & Fixed $20\%$ & 18 & 100.0 & 90.0 & 94.7 \\
              & Fixed $10\%$ & 9  & 100.0 & 45.0 & 62.1 \\
              & Fixed $30\%$ & 27 & 74.1 & 100.0 & 85.1 \\
            \addlinespace[2pt]
            \textsc{Drawer} & Auto. & 21 & 81.0 & 85.0 & 82.9 \\
              & Fixed $20\%$ & 18 & 94.4 & 85.0 & \textbf{89.5} \\
              & Fixed $10\%$ & 9  & 100.0 & 45.0 & 62.1 \\
              & Fixed $30\%$ & 27 & 66.7 & 90.0 & 76.6 \\
            \bottomrule
        \end{tabular}
    \end{minipage}\hfill
    \begin{minipage}[t]{0.49\textwidth}
        \centering
        \begin{tabular}{llcccc}
            \toprule
            Task & Rule & Rem. & Prec. & Rec. & F1 \\
            \midrule
            \textsc{Hang Cup} & Auto. & 20 & 100.0 & 100.0 & \textbf{100.0} \\
              & Fixed $20\%$ & 18 & 100.0 & 90.0 & 94.7 \\
              & Fixed $10\%$ & 9  & 100.0 & 45.0 & 62.1 \\
              & Fixed $30\%$ & 27 & 74.1 & 100.0 & 85.1 \\
            \addlinespace[2pt]
            \textsc{Tidy Pen} & Auto. & 17 & 94.1 & 80.0 & \textbf{86.5} \\
              & Fixed $20\%$ & 18 & 88.9 & 80.0 & 84.2 \\
              & Fixed $10\%$ & 9  & 100.0 & 45.0 & 62.1 \\
              & Fixed $30\%$ & 27 & 66.7 & 90.0 & 76.6 \\
            \bottomrule
        \end{tabular}
    \end{minipage}
\end{table}

We additionally evaluate the optional BIC-based procedure using only the
proficient-demonstration subsets of the four real-world tasks, corresponding
to $0\%$ contamination under our original quality-label definition. For each
task, we sample 40, 50, or 60 candidate demonstrations and repeat each size
three times, yielding nine tests per task. In 27 of the 36 tests, BIC selects
a single Gaussian and retains all candidates
(Table~\ref{tab:zero_contamination}). 

\begin{table}[H]
    \centering
    \footnotesize
    \renewcommand{\arraystretch}{1.15}
    \caption{Single-Gaussian selection at $0\%$ contamination.}
    \label{tab:zero_contamination}
    \begin{tabular}{lc}
        \toprule
        Task & Tests selecting a single Gaussian \\
        \midrule
        \textsc{Stack Cup} & 6/9 \\
        \textsc{Hang Cup} & 8/9 \\
        \textsc{Drawer} & 7/9 \\
        \textsc{Tidy Pen} & 6/9 \\
        \midrule
        Total & 27/36 ($75\%$) \\
        \bottomrule
    \end{tabular}
\end{table}

\FloatBarrier
\subsection{Qualitative Visualization}
\label{app:suboptimal_visualization}

\begin{figure}[H]
    \centering
    \captionsetup[subfigure]{font=footnotesize,skip=1pt}
    \begin{subfigure}[t]{0.42\textwidth}
        \centering
        \includegraphics[width=\linewidth]{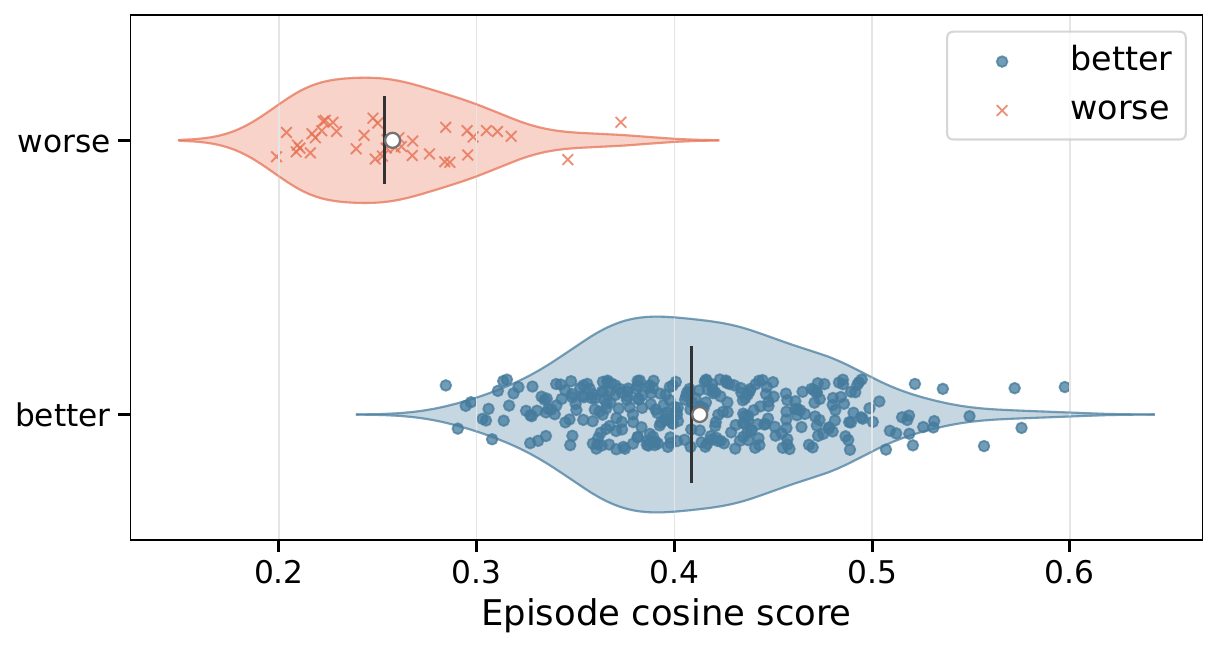}
        \caption{LIBERO-10 action corruption.}
        \label{fig:cos_libero10_action}
    \end{subfigure}\hspace{0.04\textwidth}
    \begin{subfigure}[t]{0.42\textwidth}
        \centering
        \includegraphics[width=\linewidth]{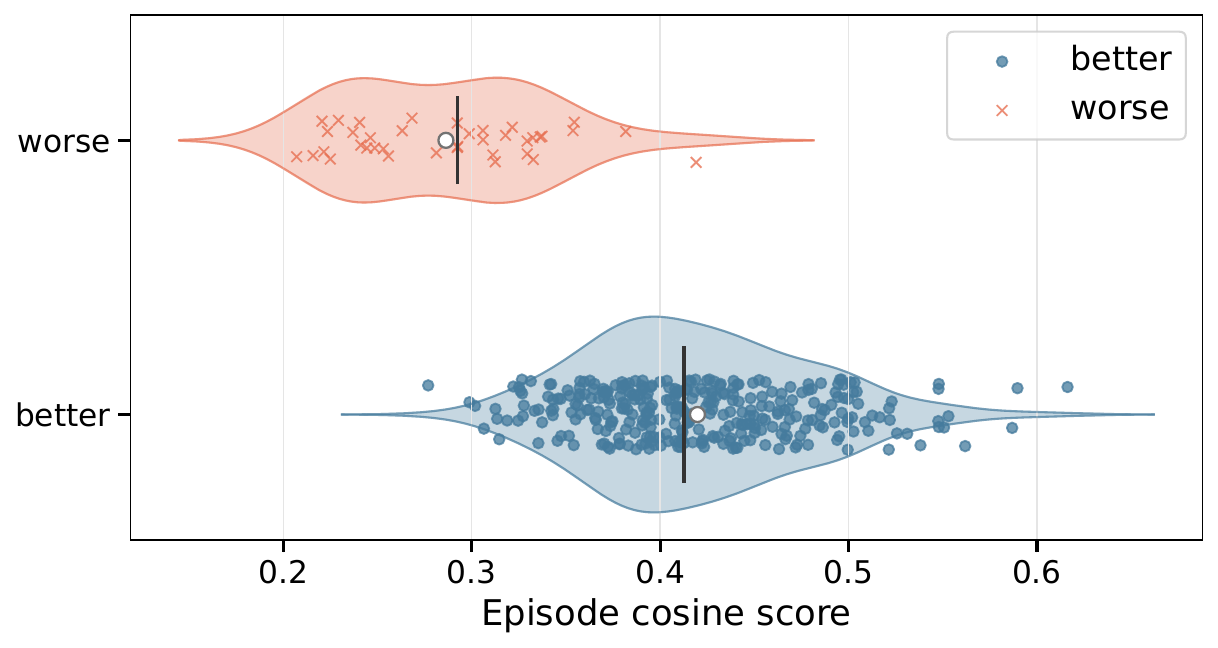}
        \caption{LIBERO-10 temporal corruption.}
        \label{fig:cos_libero10_temporal}
    \end{subfigure}

    \vspace{-3pt}
    \begin{subfigure}[t]{0.42\textwidth}
        \centering
        \includegraphics[width=\linewidth]{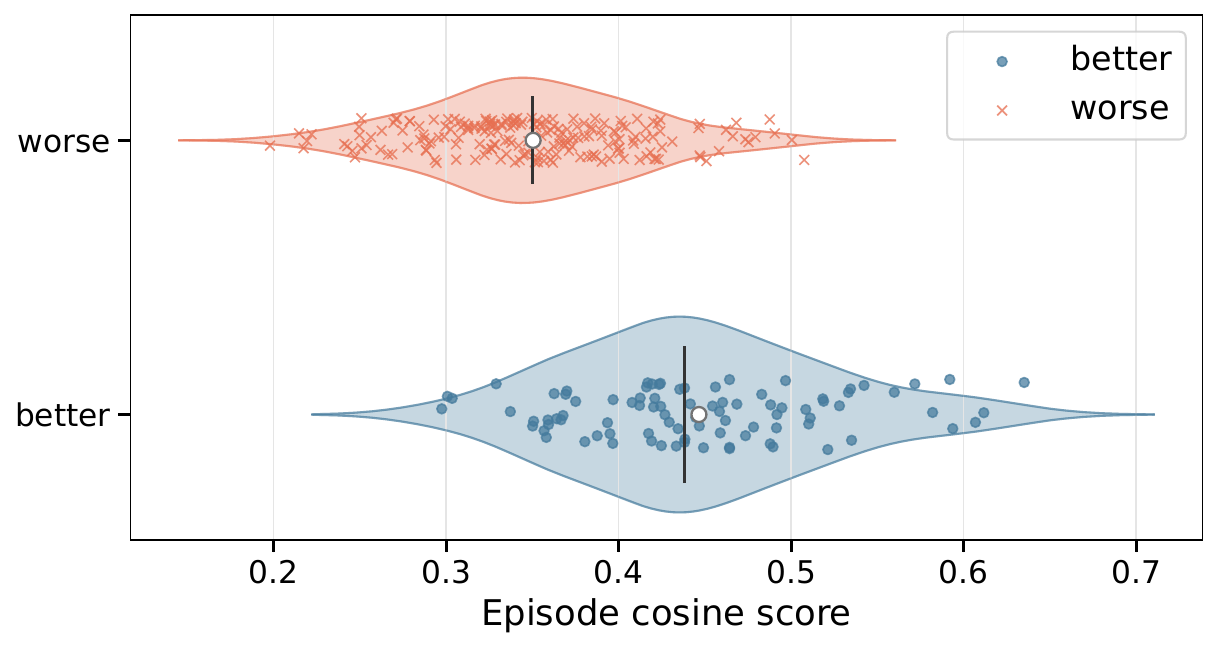}
        \caption{Robomimic-MH \textsc{Square}.}
        \label{fig:cos_robomimic_square}
    \end{subfigure}\hspace{0.04\textwidth}
    \begin{subfigure}[t]{0.42\textwidth}
        \centering
        \includegraphics[width=\linewidth]{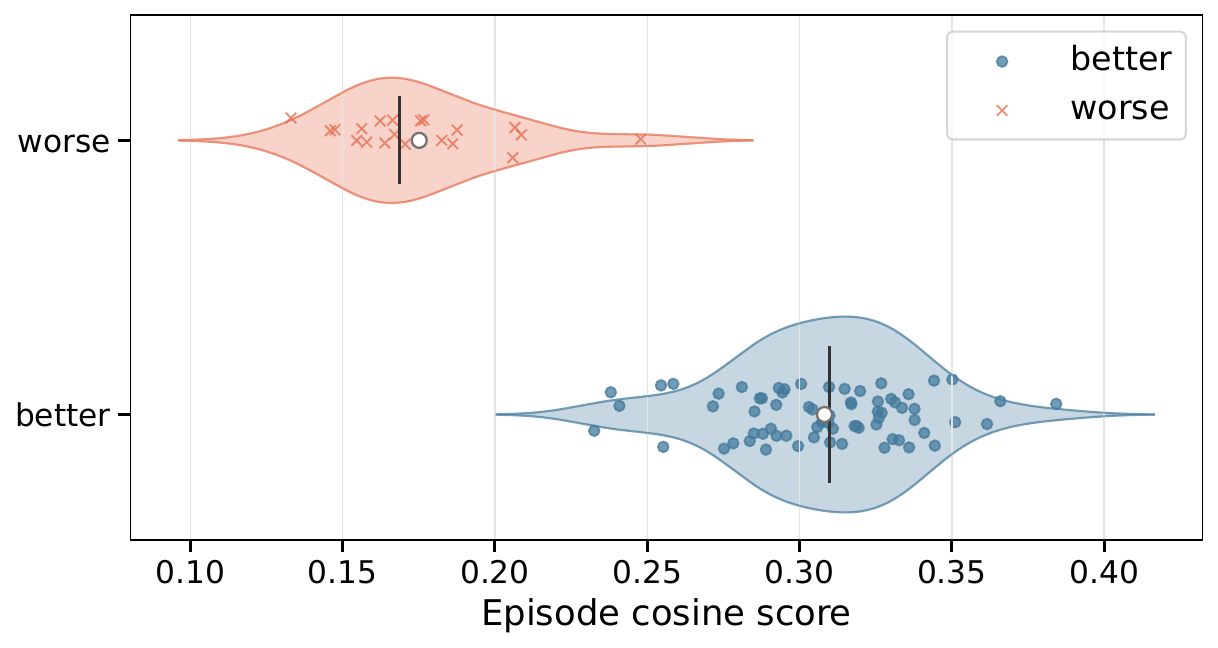}
        \caption{Real-world \textsc{Stack Cup}.}
        \label{fig:cos_realworld_stack_cup}
    \end{subfigure}
    \vspace{-2pt}
    \caption{RoboDrop episode cosine-score distributions across simulated and
    real-world data-quality settings. \emph{Worse} denotes corrupted or
    non-proficient demonstrations, and \emph{Better} denotes clean or
    proficient demonstrations.}
    \label{fig:cosine_score_visualization}
\end{figure}

Figure~\ref{fig:cosine_score_visualization} visualizes the episode-level cosine
scores produced by RoboDrop under four data-quality settings. In these plots,
\emph{Worse} denotes synthetically corrupted demonstrations or demonstrations
collected by non-proficient operators, whereas \emph{Better} denotes clean
demonstrations or those collected by proficient operators. The violin plots
show the score distributions together with the individual episode scores.

\begin{figure}[H]
    \centering
    \includegraphics[width=0.76\textwidth]{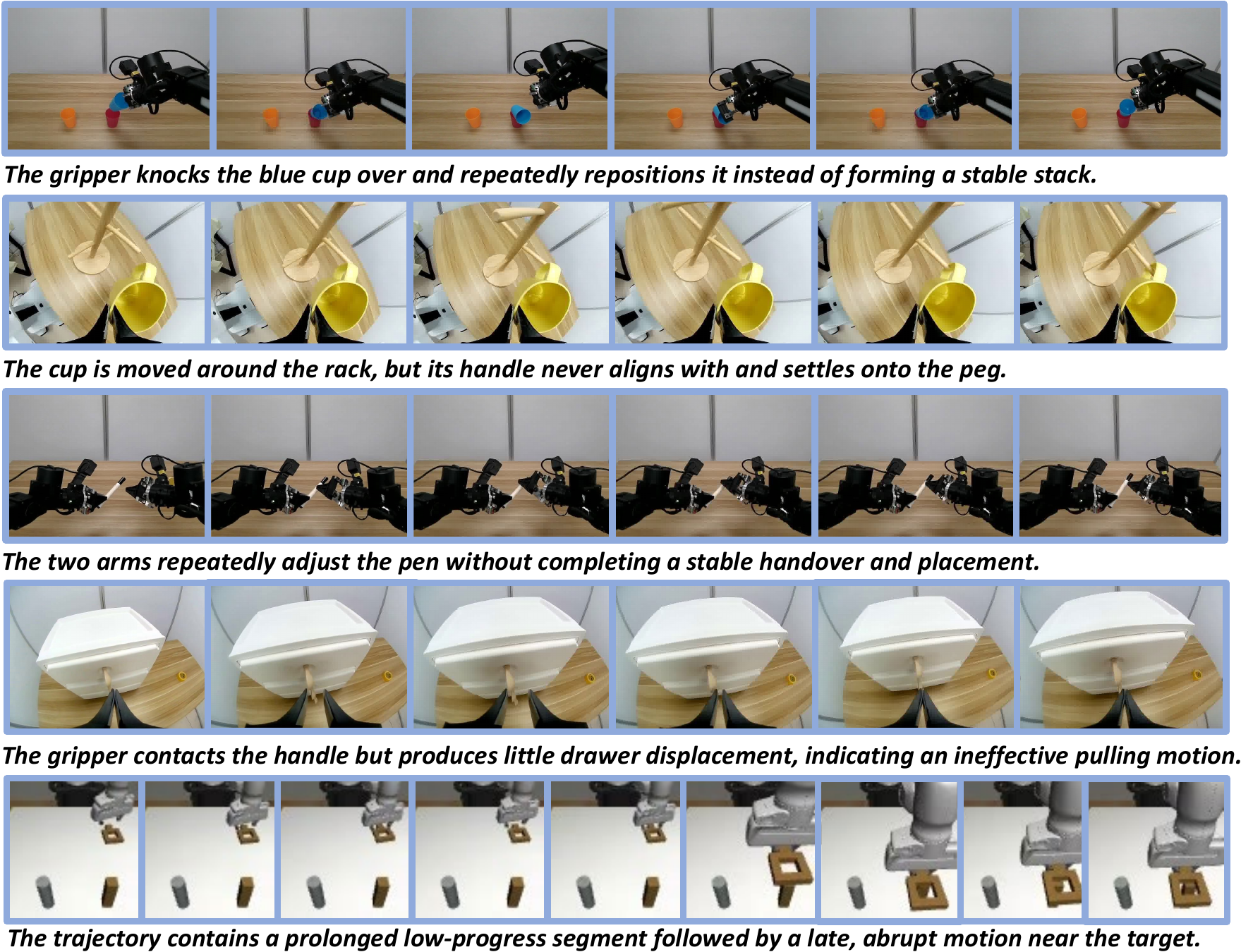}
    \caption{Representative suboptimal trajectories identified by RoboDrop.
    Each row illustrates a distinct failure mode through selected frames from
    one trajectory.}
    \label{fig:detected_suboptimal_trajectories}
\end{figure}

Figure~\ref{fig:detected_suboptimal_trajectories} presents representative low-quality trajectories identified by RoboDrop. The examples cover diverse failure modes: knocking over a cup followed by repeated corrective motions, moving a cup around the rack without aligning its handle with the peg, repeatedly adjusting a pen without completing a stable bimanual handover, contacting a drawer handle without producing meaningful displacement, and a prolonged low-progress segment followed by an abrupt late motion near the target. These cases show that RoboDrop captures locally ineffective or inconsistent behavior within a trajectory, rather than relying only on final task success or simple trajectory-level heuristics.

\clearpage
\section{Ablation Studies}
\label{app:additional_ablations}

The main text studies the effects of the local gradient reference and online
in-run scoring. Here, we examine two additional implementation choices: the
visual-neighborhood size and the subset of model parameters used for gradient
scoring. Table~\ref{tab:robodrop_ablations_detailed} reports their per-task
results.

Using only the nearest validation sample ($K_{\mathrm{vis}}=1$) achieves an
average AUROC of $86.54$, compared with $88.33$ for RoboDrop with
$K_{\mathrm{vis}}=10$. The larger neighborhood provides a more stable local
reference across tasks, although the single-neighbor setting is strongest on
\textsc{Transport}. Computing gradients through the full VLA reaches an
average AUROC of $87.02$, whereas action-expert gradients reach $88.33$ and
avoid backpropagation through the rest of the model. We therefore use
$K_{\mathrm{vis}}=10$ and action-expert gradients in the default configuration.

\begin{table}[H]
    \centering
    \footnotesize
    \renewcommand{\arraystretch}{1.12}
    \caption{Additional per-task RoboDrop ablations on Robomimic-MH with
    $\pi_{0.5}$-Base. Entries are AUROC / best balanced accuracy (\%).}
    \label{tab:robodrop_ablations_detailed}
    \begin{tabular*}{\textwidth}{@{\extracolsep{\fill}}lccccc@{}}
        \toprule
        Variant
        & \textsc{Can}
        & \textsc{Lift}
        & \textsc{Square}
        & \textsc{Transport}
        & Average \\
        \midrule
        Single neighbor ($K_{\mathrm{vis}}=1$)
        & 85.74 / 79.89
        & 87.08 / 78.33
        & 84.61 / 77.83
        & \textbf{88.72 / 84.64}
        & 86.54 / 80.17 \\

        Full-model gradients
        & 86.35 / 79.33
        & \textbf{88.90 / 82.00}
        & 88.19 / 83.03
        & 84.63 / 79.58
        & 87.02 / 80.99 \\

        \textbf{RoboDrop}
        & \textbf{88.88 / 82.75}
        & 87.23 / 79.44
        & \textbf{89.07 / 83.14}
        & 88.15 / 81.97
        & \textbf{88.33 / 81.83} \\
        \bottomrule
    \end{tabular*}
\end{table}

\paragraph{CountSketch dimension.}
Per-sample action-expert gradients are high-dimensional, so RoboDrop applies
CountSketch before computing gradient compatibility. We evaluate target
dimensions of 1024, 2048, and 4096 on the four Robomimic-MH tasks. As shown in
Table~\ref{tab:countsketch_dimension}, performance is stable across compression
levels, while 4096 dimensions achieve the strongest mean AUROC and balanced
accuracy ($88.33\%$ and $81.83\%$, respectively). We therefore use 4096 as the
default, providing substantial gradient compression without sacrificing the
quality signal used for episode ranking.

\begin{table}[H]
    \centering
    \footnotesize
    \renewcommand{\arraystretch}{1.12}
    \caption{Sensitivity to the CountSketch target dimension on Robomimic-MH
    with $\pi_{0.5}$-Base (\%). Best bacc denotes best balanced accuracy.}
    \label{tab:countsketch_dimension}
    \begin{tabular*}{\textwidth}{@{\extracolsep{\fill}}lcccccccc@{}}
        \toprule
        & \multicolumn{2}{c}{\textsc{Can}}
        & \multicolumn{2}{c}{\textsc{Lift}}
        & \multicolumn{2}{c}{\textsc{Square}}
        & \multicolumn{2}{c}{\textsc{Transport}} \\
        \cmidrule(lr){2-3}\cmidrule(lr){4-5}
        \cmidrule(lr){6-7}\cmidrule(lr){8-9}
        Dimension
        & AUROC & Best bacc
        & AUROC & Best bacc
        & AUROC & Best bacc
        & AUROC & Best bacc \\
        \midrule
        1024
        & 86.62 & 79.07
        & 88.40 & \textbf{80.51}
        & 85.35 & 79.52
        & 87.78 & \textbf{82.39} \\
        2048
        & 86.37 & 79.26
        & \textbf{88.71} & 80.01
        & 84.56 & 78.21
        & \textbf{88.22} & 82.33 \\
        \textbf{4096}
        & \textbf{88.88} & \textbf{82.75}
        & 87.23 & 79.44
        & \textbf{89.07} & \textbf{83.14}
        & 88.15 & 81.97 \\
        \bottomrule
    \end{tabular*}
\end{table}

\paragraph{Validation-gradient refresh interval.}
RoboDrop refreshes the validation reference gradients every $M$ optimization
steps during warm-up. We vary $M$ on the Robomimic-MH \textsc{Square} task
while holding all other settings fixed. Table~\ref{tab:validation_refresh_interval}
shows that $M=200$ achieves the best AUROC and balanced accuracy, while
performance generally declines when the interval is increased beyond 400
steps. Refreshing every 100 steps performs comparably but does not improve over
$M=200$. These results support the default interval of $M=200$ and indicate
that overly stale validation gradients weaken the compatibility signal.

\begin{table}[H]
    \centering
    \footnotesize
    \setlength{\tabcolsep}{9pt}
    \renewcommand{\arraystretch}{1.10}
    \caption{Sensitivity to the validation-gradient refresh interval $M$ on
    Robomimic-MH \textsc{Square} with $\pi_{0.5}$-Base (\%).}
    \label{tab:validation_refresh_interval}
    \begin{tabular}{ccc}
        \toprule
        Refresh interval $M$ & AUROC & Best bacc \\
        \midrule
        100  & 88.17 & 82.70 \\
        \textbf{200} & \textbf{89.07} & \textbf{83.14} \\
        400  & 85.93 & 79.26 \\
        600  & 84.36 & 77.08 \\
        800  & 77.83 & 72.70 \\
        1000 & 78.76 & 73.32 \\
        \bottomrule
    \end{tabular}
\end{table}

\clearpage
\subsection{DINO Similarity and Cross-Episode Gradient Alignment}
\label{app:dino_gradient_alignment}

\begin{wrapfigure}{r}{0.50\textwidth}
    \vspace{-8pt}
    \centering
    \includegraphics[width=0.47\textwidth]{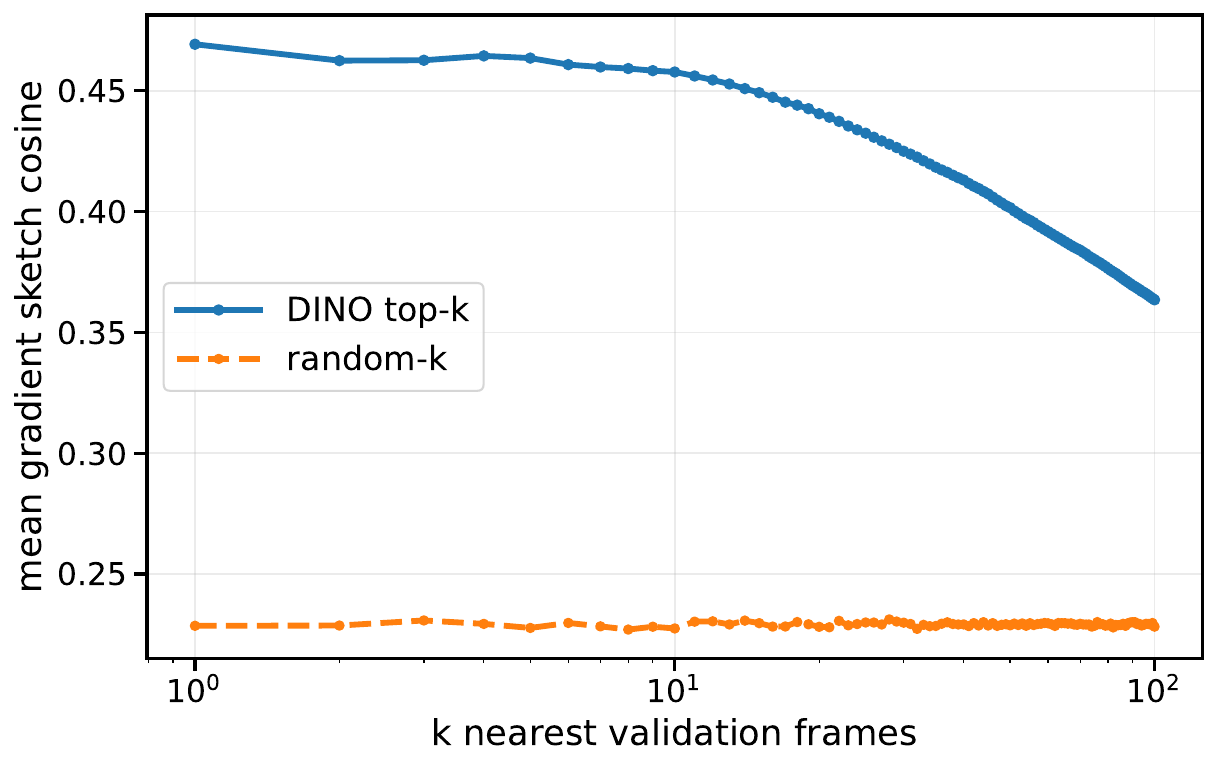}
    \caption{Cross-episode gradient-direction similarity for DINO-nearest and
    randomly sampled frames on Robomimic-MH \textsc{Square}.}
    \label{fig:gradient_alignment}
    \vspace{-10pt}
\end{wrapfigure}

We validate DINO similarity as the visual retrieval signal using
$\pi_{0.5}$-Base on the \textsc{Square} task from Robomimic-MH. At a fixed
post-training checkpoint, we compute action-expert gradients for individual
frames using the same model parameters. To prevent temporally adjacent frames
from trivially matching, all comparisons are cross-episode: for each query
frame, we exclude its source episode and rank frames from the remaining
episodes by cosine similarity between their frozen DINO features. For each
neighborhood size $k$, we then compare the average gradient-direction cosine of
the $k$ DINO-nearest frames with that of $k$ randomly sampled cross-episode
frames.

As shown in Figure~\ref{fig:gradient_alignment}, DINO-nearest frames have
substantially higher gradient-direction similarity than random cross-episode
frames. Their alignment gradually decreases as $k$ grows and progressively less
similar frames enter the neighborhood, while the random baseline remains low.
This relationship indicates that visually similar manipulation states tend to
support similar update directions and therefore justifies DINO similarity for
constructing the local validation reference.

\subsection{Detecting Localized Errors}
\label{app:localized_corruption}
To test whether episode averaging obscures errors confined to a critical stage,
we construct a localized temporal-corruption setting on LIBERO-Object. A
vision--language model identifies grasping events, and temporal corruption is
restricted to a four-second window extending two seconds before and after each
event. With its default full-trajectory mean aggregation, RoboDrop achieves
$85.90\%$ AUROC, outperforming other baselines
(Table~\ref{tab:localized_corruption_baselines}). This shows that episode-level
averaging can retain a useful signal even when errors affect only a short
segment.

\begin{table}[!htbp]
    \centering
    \footnotesize
    \caption{Episode-level AUROC (\%) under localized temporal corruption on
    LIBERO-Object.}
    \label{tab:localized_corruption_baselines}
    \begin{tabular*}{\textwidth}{@{\extracolsep{\fill}}lccccc@{}}
        \toprule
        Method & Behavior Retrieval & DataMIL & Scizor & QoQ & RoboDrop \\
        \midrule
        AUROC $\uparrow$ & 67.59 & 64.85 & 54.62 & 84.46 & 85.90 \\
        \bottomrule
    \end{tabular*}
\end{table}

\FloatBarrier
\subsection{Episode-Level Aggregation}
\label{app:episode_aggregation}
Using the localized-corruption setting in Appendix~\ref{app:localized_corruption},
we assess sensitivity to the aggregation rule by comparing the
full-trajectory mean with the mean of the lowest-scoring $10\%$ or $20\%$ of
frames and the lowest mean over a contiguous one-second window. These rules
aggregate the same frame-level scores into episode-level scores.

\begin{table}[!htbp]
    \centering
    \footnotesize
    \caption{Episode-level AUROC (\%) for alternative aggregation rules under
    localized temporal corruption on LIBERO-Object.}
    \label{tab:localized_aggregation}
    \begin{tabular}{lcc}
        \toprule
        Aggregation &  RoboDrop \\
        \midrule
        Full-trajectory mean &  85.90 \\
        Lowest-scoring $10\%$ mean &  77.43 \\
        Lowest-scoring $20\%$ mean &  83.95 \\
        Worst contiguous 1-second window mean  & \textbf{88.70} \\
        \bottomrule
    \end{tabular}
\end{table}

As shown in Table~\ref{tab:localized_aggregation}, the worst-window rule
improves RoboDrop's AUROC by $2.80$ percentage points over full-trajectory
averaging. However, this advantage does not carry over to naturally collected
Robomimic-MH data: full-trajectory averaging achieves the highest AUROC on all
four tasks (Table~\ref{tab:natural_aggregation}). These results support using
the mean as the default when the error pattern is unknown, while contiguous
window aggregation can improve sensitivity to localized corruption.

\begin{table}[!htbp]
    \centering
    \footnotesize
    \caption{RoboDrop episode-level AUROC (\%) under different aggregation rules
    on naturally collected Robomimic-MH demonstrations.}
    \label{tab:natural_aggregation}
    \begin{tabular}{lcccc}
        \toprule
        Task & Full mean & Lowest $10\%$ & Lowest $20\%$ & Worst 1-second window \\
        \midrule
        \textsc{Can} & \textbf{88.88} & 72.52 & 78.25 & 86.42 \\
        \textsc{Lift} & \textbf{87.23} & 67.29 & 72.42 & 85.26 \\
        \textsc{Square} & \textbf{89.07} & 68.78 & 74.44 & 78.61 \\
        \textsc{Transport} & \textbf{88.15} & 73.46 & 77.54 & 69.90 \\
        \midrule
        Average & \textbf{88.33} & 70.51 & 75.66 & 80.05 \\
        \bottomrule
    \end{tabular}
\end{table}

\FloatBarrier
\section{Validation Loss and Rollout Success During Early Post-Training}
\label{app:validation-loss}

We examined how validation flow-matching loss and rollout success vary during
early post-training on Robomimic Can and Lift. For each task, we evaluated
five checkpoints spanning approximately the first 1.5 epochs. We ran
100 rollouts per checkpoint with matched evaluation seeds within each task.

\begin{table}[H]
\centering
\caption{Validation flow-matching loss and rollout success during early
post-training on Robomimic Can and Lift.}
\label{tab:early-validation-loss}
\begin{tabular}{rrr|rrr}
\toprule
\multicolumn{3}{c|}{Can} & \multicolumn{3}{c}{Lift} \\
Step & Val. loss & Success (\%) & Step & Val. loss & Success (\%) \\
\midrule
500   & 2.157 & 0  & 400   & 1.216 & 0  \\
1,000 & 1.530 & 0  & 600   & 1.122 & 0  \\
1,500 & 0.846 & 11 & 800   & 1.163 & 1  \\
2,000 & 0.787 & 12 & 1,000 & 0.958 & 53 \\
2,500 & 0.615 & 43 & 1,200 & 0.607 & 91 \\
\bottomrule
\end{tabular}
\end{table}

As shown in Table~\ref{tab:early-validation-loss}, validation flow-matching loss
and rollout success exhibit a clear negative association across the early
checkpoints on both tasks. This observation concerns success at the checkpoints
evaluated but leaves open whether a larger early reduction in loss leads to
higher success after convergence. Although RoboDrop uses early validation loss as a surrogate for data selection, it achieves considerable improvement on downstream success rate. As the scoring framework can accommodate other differentiable reference objectives, we leave the exploration of better surrogates for success rate to future work.

\section{Training and Evaluation Details}
\label{app:training_details}

\paragraph{Warm-up Training.}
Unless otherwise specified, experiments use the open-source
$\pi_{0.5}$-Base model. RoboDrop performs one warm-up epoch over the complete
candidate set. Candidate samples are shuffled before warm-up and each is scored
exactly once, at the step when it enters a minibatch. For each candidate sample,
RoboDrop retrieves $K_{\mathrm{vis}}=10$ validation samples after task-semantic
matching and uses their most recently cached reference gradients. The validation
gradients are refreshed every $M=200$ optimization steps and reused between
refreshes; no additional candidate-set scoring sweep is performed. Only the
action-expert parameters participate in per-sample gradient scoring. Each
gradient is compressed to 4096 dimensions with CountSketch before local cosine
compatibility is computed and stored for episode-level aggregation. Warm-up
training uses four NVIDIA A100 GPUs with a global batch size of 32. A one-epoch
warm-up run takes approximately 2 hours on each of LIBERO-Spatial,
LIBERO-Object, and LIBERO-Goal, 4 hours on LIBERO-10, 5.4 hours on average
across the four Robomimic tasks and 4.5 hours on average across the four real-robot tasks. 

\paragraph{Post-training.}
Unless a fixed removal budget is specified, RoboDrop uses BIC to select between the single-Gaussian and two-component score models described in Section~\ref{sec:gmm_filtering}. It retains every candidate
episode when the single Gaussian is selected. When the two-component model is
selected, episodes whose posterior probability of belonging to the lower-mean
component exceeds $q=0.8$ are removed. The warm-up policy is then discarded,
and a fresh pretrained policy is post-trained on the retained
candidate episodes together with the clean validation episodes. 

For LIBERO, we fully fine-tune all parameters of $\pi_{0.5}$-Base with a global batch size of
32. Post-training runs for 20,000 optimization steps on each of LIBERO-Object,
LIBERO-Spatial, and LIBERO-Goal, and for 30,000 steps on LIBERO-10. For Robomimic-MH, we use the open-source X-VLA-Base model and remove the lowest-scoring $50\%$ of the candidate episodes for all selection methods. We fully fine-tune all model parameters with a global batch size of 32 for 10,000 optimization steps on \textsc{Can}, \textsc{Lift}, and \textsc{Square}, and for 30,000 steps on \textsc{Transport}. For each Robomimic-MH task and curation method, we repeat post-training and rollout evaluation over three runs with different training and evaluation seeds, and report the mean success rate across runs. For each real-world task, post-training uses a global batch size of 32 and runs for 30,000 optimization steps.

\paragraph{Evaluation.}
We evaluate episode-level scores using AUROC and best balanced accuracy.
AUROC measures ranking quality across decision thresholds, with $50\%$
corresponding to chance-level discrimination and $100\%$ to perfect separation.
Balanced accuracy averages the true-positive and true-negative rates, giving
equal weight to both classes:
\[
    \operatorname{BAcc}(t)
    = \frac{\operatorname{TPR}(t)+\operatorname{TNR}(t)}{2}.
\]
We report the best balanced accuracy over all score thresholds,
$\max_t \operatorname{BAcc}(t)$, to characterize achievable class separation.
For synthetic corruptions, evaluation labels indicate whether an episode was
corrupted; for naturally collected demonstrations, operator proficiency
provides a proxy label for quality.  We measure policy success as the fraction of successful rollouts.
For LIBERO, each policy is evaluated over 20 rollouts per task using the same
evaluation seeds across methods. For Robomimic-MH, each method is post-trained
with three different training seeds, and each resulting policy is evaluated using 3 eval seeds
over 100 rollouts per task for 3 times. For the real-world experiments, each policy is evaluated over 20 rollouts per
task. Suite-level and cross-task results are computed by averaging task-level
success rates.

\end{document}